\pdfoutput=1

\documentclass[11pt]{article}

\usepackage[preprint]{acl}

\usepackage{times}
\usepackage{latexsym}

\usepackage[T1]{fontenc}

\usepackage[utf8]{inputenc}

\usepackage{microtype}

\usepackage{inconsolata}

\usepackage{graphicx}
\usepackage{booktabs}

\usepackage{todonotes}
\usepackage{amsmath}

\usepackage{booktabs}
\usepackage{hyperref}
\usepackage{url}
\usepackage[most]{tcolorbox}
\usepackage{xcolor}
\usepackage{lipsum}
\usepackage{hyperref} 
\title{Pruning Weights but Not Truth: Safeguarding Truthfulness \\ While Pruning LLMs}

\author{
  Yao Fu\textsuperscript{1}, 
  Runchao Li\textsuperscript{1},
  Xianxuan Long\textsuperscript{1}, \\
  \textbf{Haotian Yu\textsuperscript{1},}
  \textbf{Xiaotian Han\textsuperscript{1},}
  \textbf{Yu Yin\textsuperscript{1},}
  \textbf{Pan Li\textsuperscript{2}\footnotemark[1]} \\
  \textsuperscript{1}Case Western Reserve University \\
  \textsuperscript{2}Hangzhou Dianzi University \\
  \texttt{\{yxf484,rxl685,xxl1514,hxy692,xxh584,yxf1421\}@case.edu}, \\
  \texttt{lipan@ieee.org}
}

\begin{document}
\maketitle
\begin{abstract}

Neural network pruning has emerged as a promising approach for deploying LLMs in low-resource scenarios while preserving downstream task performance. However, for the first time, we reveal that such pruning disrupts LLMs' internal activation features crucial for lie detection, where probing classifiers (typically small logistic regression models) trained on these features assess the truthfulness of LLM-generated statements. This discovery raises a crucial open question: how can we prune LLMs without sacrificing these critical lie detection capabilities?  Our investigation further reveals that naively adjusting layer-wise pruning sparsity based on importance inadvertently removes crucial weights, failing to improve lie detection performance despite its reliance on the most crucial LLM layer. To address this issue, we propose \textbf{T}ruthful \textbf{P}runing aligned by \textbf{L}ayer-wise \textbf{O}utliers (\textbf{TPLO}), which places greater emphasis on layers with more activation outliers and stronger discriminative features simultaneously. This preserves LLMs' original performance while retaining critical features of inner states needed for robust lie detection. Moreover, we introduce a prompting rule to enrich the TruthfulQA benchmark for better calibrating LLM pruning. Empirical results show that our approach improves the hallucination detection \footnote{In this paper, we adopt the definition of "hallucinations" from recent works \cite{bayat2024enhanced, zhang2024truthx}, referring to instances where LLMs produce fluent, instruction-compliant, yet untruthful responses. Consequently, we use the terms "lie detection" and "hallucination detection" interchangeably throughout this work.} for pruned LLMs (achieving 88\% accuracy at 50\% sparsity) and enhances their performance on TruthfulQA. Codes and data are available \href{https://github.com/ClarkFu007/TPLO}{here}.

\end{abstract}

\begin{figure}[t]
  \includegraphics[width=\columnwidth]{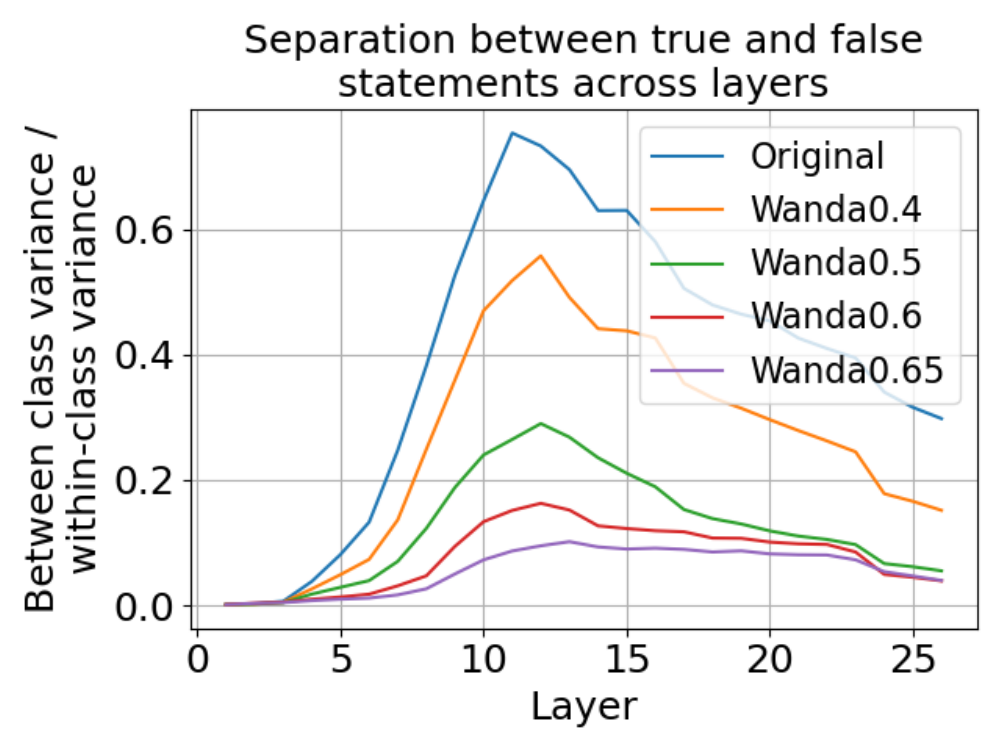}
  \caption{Each curve represents the layer-wise ratio of between-class variance to within-class variance for activations corresponding to true and false statements. This ratio is averaged across all dimensions within each LLM layer, indicating that layers with a higher ratio contain more discriminative features for distinguishing between true and false statements, whereas layers with a lower ratio have fewer. We define this metric as \textbf{L}ayer-wise \textbf{S}eparability of True and False \textbf{D}istribution (\textbf{LSD}). Three key takeaways: i) Original models (unpruned LLaMA3.1-8B-Instruct) have the best ability to separate true/false statements. ii) Moderate pruning (e.g., less than 0.5 sparsity) retains reasonable performance, but heavy pruning (e.g., 0.65 sparsity) significantly degrades separation ability. iii) The most useful layers for classifying true/false statements seem to be consistently around layer 10-15 no matter what sparsity is.}
  \label{fig:cities_comparison}
\end{figure}

\section{Introduction}
\label{section:Introduction}
Large language models (LLMs) \cite{llm_survey} are remarkably impressive across a wide range of natural language processing (NLP) tasks \cite{llm_nlp_survey}. Despite their potential usefulness, the substantial computational and memory requirements of LLM inference pose challenges for deployment in resource-constrained scenarios \cite{llm_challenge_survey}. Consequently, there has been a surge of interest in how to effectively convert LLMs into compact ones for reducing storage and accelerating inference \cite{llm_compression_survey}. Network pruning \cite{han_pruning}, one of the most representative approaches in model compression, demonstrates the possibility of removing around 50\% of LLMs' active parameters \cite{wanda}, or even more \cite{owl} with minimal performance degradation. However, most pruning techniques primarily focus on ensuring that the compressed LLMs have low perplexity and good performance on some zero-shot tasks \cite{eval-harness, owl}, which is far from thoroughly understanding the generalization of LLMs after being pruned.

A recent fascinating discovery \cite{burger2024truth} motivating our work, is that not only can LLMs be instructed to "lie" (defined as hallucination where they knowingly generate false statements) but also they can engage in strategic deception to achieve specific goals, even for models trained to be honest. Experimenting with various LLMs (including LLaMA \cite{llama2, llama3} and Mistral \cite{mistral7b} model families), the authors \cite{burger2024truth} find out a global truth direction $t_G$ that generalizes across a broad spectrum of true/false statement types beyond the training set, which brings the possibility of general-purpose lie detection of LLMs. Yet the LLMs used in \citet{burger2024truth} are all original and uncompressed. \textbf{This raises a central question:} \textit{Can we enable robust lie detection in compressed LLMs deployed on edge devices, allowing users to diagnose whether models are knowingly generating falsehoods under limited computational resources?}

We study this question by training classifiers on the internal activations of pruned LLMs to judge whether a given statement is true or false, using both supervised \cite{azaria2023internal, williams2023does} and unsupervised techniques \cite{burns2022discovering}. We discover that after being unstructuredly pruned via Wanda \cite{wanda} at sparsity of 50\% followed with \citet{bandari2024c4}, the quality of LLMs' internal activations will deteriorate, leading to a less robust lie detector shown in Figure \ref{fig:bar_comparison_original_wanda0.5}. We conjecture that applying a uniform pruning ratio across all layers, where each layer is pruned at the same sparsity, is detrimental to the training of robust lie detectors, as intermediate activation features might contribute differently to lie detection across layers. To validate this, we visualize the LSD of the original LLaMA3.1-8B-Instruct in Figure \ref{fig:cities_comparison} and observe that each LLM layer exhibits varying degrees of discriminative quality in its internal states, suggesting that some layers are more effective at distinguishing between true and false statements while others are not. Based on this observation, we construct a baseline method, \textbf{S}eparability \textbf{W}eighted \textbf{L}ayer-wise sparsity (\textbf{SWL}), which adjusts each layer's pruning sparsity inversely proportional to its layer-wise separability.

However, simply applying SWL may inadvertently prune more important weights, as suggested by OWL \cite{owl} which highlights that LLM outlier distributions across layers follow a distinctly non-uniform pattern that does not fully align with LSD. Recognizing OWL as a valuable indicator for effectively optimizing layer-wise sparsity strategies in LLM pruning, we propose \textbf{T}ruthful \textbf{P}runing aligned by \textbf{L}ayer-wise \textbf{O}utliers (\textbf{TPLO}). TPLO enhances layer-wise pruning by aligning LSD with outlier ratio distributions, ensuring a more effective sparsity allocation. Our approach is based on the insight that greater emphasis should be synchronously placed on layers with a higher prevalence of outliers and more discriminative activation vectors, ensuring that the original performance of LLMs is maintained while preserving more internal features essential for training lie detectors. Furthermore, we propose a novel prompting rule to enrich the TruthfulQA benchmark \cite{truthful_qa} as extra calibration data to help prune LLMs inspired from \citet{bandari2024c4} that C4, one of data sources to pretrain LLMs, is not the optimal choice for calibrating LLM pruning. In summary, the contributions of this paper are as follows: 
\begin{itemize}
    \item We conduct an in-depth investigation into how pruning influences LLMs' internal states that are used for training robust lie detectors.
    \item We propose a method called \textbf{T}ruthful \textbf{P}runing aligned by \textbf{L}ayer-wise \textbf{O}utliers (\textbf{TPLO}) to enhance lie detection in pruned LLMs.
    \item  We introduce a prompt rule to enrich the TruthfulQA benchmark \cite{truthful_qa} to help calibrate truthful LLM pruning.
\end{itemize}

\section{Related Work}
\label{section:Related Work}
\subsection{LLM Pruning}
Network pruning is a popular method to reduce sizes of neural networks (NNs) with minimal performance loss \cite{han_pruning}. Many pruning techniques developed for computer vision (CV) heavily relies on adequate retraining \cite{pruning_survey}, indicating that pruning LLMs is inaccessible for practitioners with limited computational resources. Wanda \cite{wanda} is the first work to prune LLMs without updating weights that prunes parameters based on the product of their magnitude and input activations. OWL \cite{owl} presents a novel approach to non-uniformly pruning LLMs in terms of outlier score distributions. The methods above are \textit{\textbf{unstructured pruning}}, which targets individual parameters and creates an irregular sparsity pattern. \textit{\textbf{Structured pruning}} \cite{zhang2023loraprune, wei2024structured, gao2024disp, slimgpt} uses larger units such as rows or columns of weight matrices, while \textit{\textbf{semi-structured pruning}} \cite{sparsellm, maskllm} designs N:M patterns (N elements are non-zero for every M consecutive ones) to facilitate faster inference supported by specialized hardware \cite{pool2021accelerating}. Our work \textbf{differs} from prior studies in that we propose a pruning method that preserves the internal features necessary for lie detection, enabling us to diagnose whether pruned models still "know" they are lying, instead of aiming to outperform existing pruning methods.

\begin{figure}[t]
  \includegraphics[width=\columnwidth]{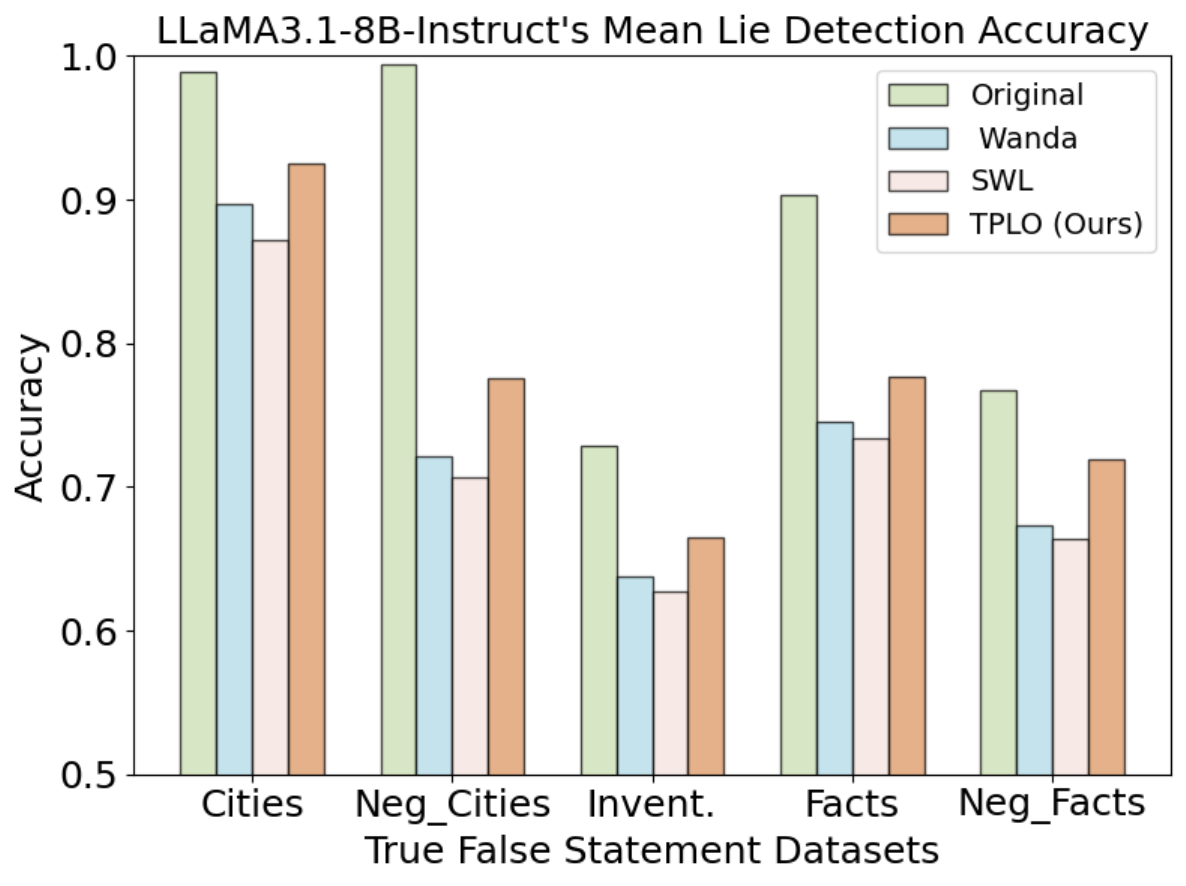}
  \caption{The visualization of the impact of 50\% sparsity (via Wanda, SWL, and TPLO) on LLaMA3.1-8B-Instruct's probing (lie detection) accuracy across several true false datasets via logistic regression.}
  \label{fig:bar_comparison_original_wanda0.5}
\end{figure}

\subsection{Calibration and Evaluation Study in LLM Compression}
Existing LLM pruning methods use C4 (128 samples, 2048 tokens), a large-scale and cleaned web text dataset derived from Common Crawl \cite{common_crawl}, as the default calibration set to compute pruning scores \cite{wanda, owl, wei2024structured}. The first comprehensive study of how different types of calibration data affect the performance of pruned or quantized LLMs is conducted by \citet{williams2023does}, which is limited to pretraining data sources. Expanding on this, \citet{bandari2024c4} analyze not only four widely used pretraining datasets but also a diverse set of downstream datasets, which incorporates In-Context Learning (ICL) \cite{icl_survey} and Chain of Thought (CoT) prompting \cite{cot}. Recently, \citet{ji2024beware} introduce a self-generating calibration data synthesis strategy to construct more effective calibration datasets. \textbf{In this work}, we investigate whether combining C4 with the enriched TruthfulQA data could help better detect whether pruned LLMs are "lying".

\subsection{Lie Detection in LLMs}
As LLMs become increasingly widespread, robustly detecting when they lie is an important research topic. \citet{pacchiardi2023catch} propose a black-box lie detection method that relies only on model outputs. In contrast, other studies use internal activations to discern truthfulness, using both supervised \cite{azaria2023internal, li2024inference} and unsupervised \cite{burns2022discovering} techniques. Notably, both \citet{azaria2023internal} and \citet{marks2023geometry} identify a linear "truth direction" in activation space that separates true from false statements. Interestingly, \citet{azaria2023internal} show that classifiers trained on both affirmative and negated statements generalize across topics, while those trained only on affirmatives fail to generalize to negated ones. \citet{burger2024truth} further explains this by revealing a two-dimensional subspace where true and false statements are separable. \textbf{In this work}, we use datasets from \citet{burger2024truth} to analyze whether pruned LLMs are more prone to "lying" or not. Unlike recent studies that systematically evaluate how compression affects LLMs' safety dimensions (bias, toxicity, and fairness) \cite{decoding_compressed_trust, beyond_perplexity}, we study how does pruning influence LLMs' internal states used for detecting lies and propose mitigation techniques. Moreover, we demonstrate that DoLa \cite{dola}, an orthogonal self-decoding strategy designed to reduce hallucinations during inference, can be \textbf{seamlessly integrated} into our framework to further enhance the truthfulness of pruned LLMs' responses.

\section{How Does Pruning Influence LLMs' Lie Detection?}
In this section, we provide a detailed analysis of how pruning impacts LLMs' inner states that are used to train lie (hallucination) detectors. 

\subsection{Setup}
\label{subsection:setup}
\paragraph{Models.} We follow the recent LLM lie-detection work \cite{burger2024truth} to use LLaMA3.1-8B-Instruct \cite{llama3} \footnote{\url{https://huggingface.co/meta-llama/Llama-3.1-8B-Instruct}} in the main text to do experiments. In Appendix \ref{appendix: Results for Other LLMs}, we demonstrate that similar experimental results appear in LLaMA2-13B-chat \cite{llama2} \footnote{\url{https://huggingface.co/meta-llama/Llama-2-13b-chat-hf}} and Mistral-7B-Instruct-v0.3 \cite{mistral7b} \footnote{\url{https://huggingface.co/mistralai/Mistral-7B-Instruct-v0.3}}.

\paragraph{Pruning Methods.} We choose Wanda \cite{wanda} to study because it is the first LLM pruning technique that requires no retraining or weight updates, making it highly suitable for organizations with limited computational resources. Specifically, we consider the input feature activations of a layer as $\mathbf{X}$ with dimensions $(N \times L, C_{\text{in}})$, where $N$ and $L$ represent the batch size and sequence dimensions, respectively. The weight matrix $\mathbf{W}$ has dimensions $(C_{\text{out}}, C_{\text{in}})$, where $C_{\text{in}}$ and $C_{\text{out}}$ represent the number of input and output channels, respectively. The \textbf{weight importance score} is computed as:
\[
\mathbf{A}_{i,j} = \|\mathbf{X}_j\|_2 \cdot |\mathbf{W}_{i,j}|,
\]
which is the aggregation of all input activations connected to weight $\mathbf{W}_{i,j}$, multiplied by its magnitude $|\mathbf{W}_{i,j}|$. Here, $\|\mathbf{X}_j\|_2$ is the $\ell_2$ norm of the $j$-th feature of input $\mathbf{X}$. 
This computation is performed across all $N \times L$ tokens, resulting in a scalar value denoted as $\|\mathbf{X}_j\|_2$.
For each layer, weights with relatively low scores will be pruned (set as 0) at a given pruning sparsity, resulting in sparse LLMs. Following \citet{bandari2024c4}, our paper focuses on unstructured pruning at sparsity of 50\%.

\paragraph{True-False Datasets.} To explore the internal truth representation of pruned LLMs, we borrow several public labeled datasets of true and false English statements from the recent work \cite{burger2024truth}, which consists of 6 different topics: "animal\_class", "cities", "inventors", "element\_symb", "facts", and "sp\_en\_trans", as well as 2 different grammatical structures: affirmative statements and negated statements. Affirmative statements are structured similarly to the context statement examples in the original true/false dataset, while negated statements are formed by negating the affirmative statements using the word "not." More detailed introduction of true/false datasets is in Appendix \ref{appendix:Details of True False Dataset}.

\begin{figure}[t]
  \includegraphics[width=\columnwidth]{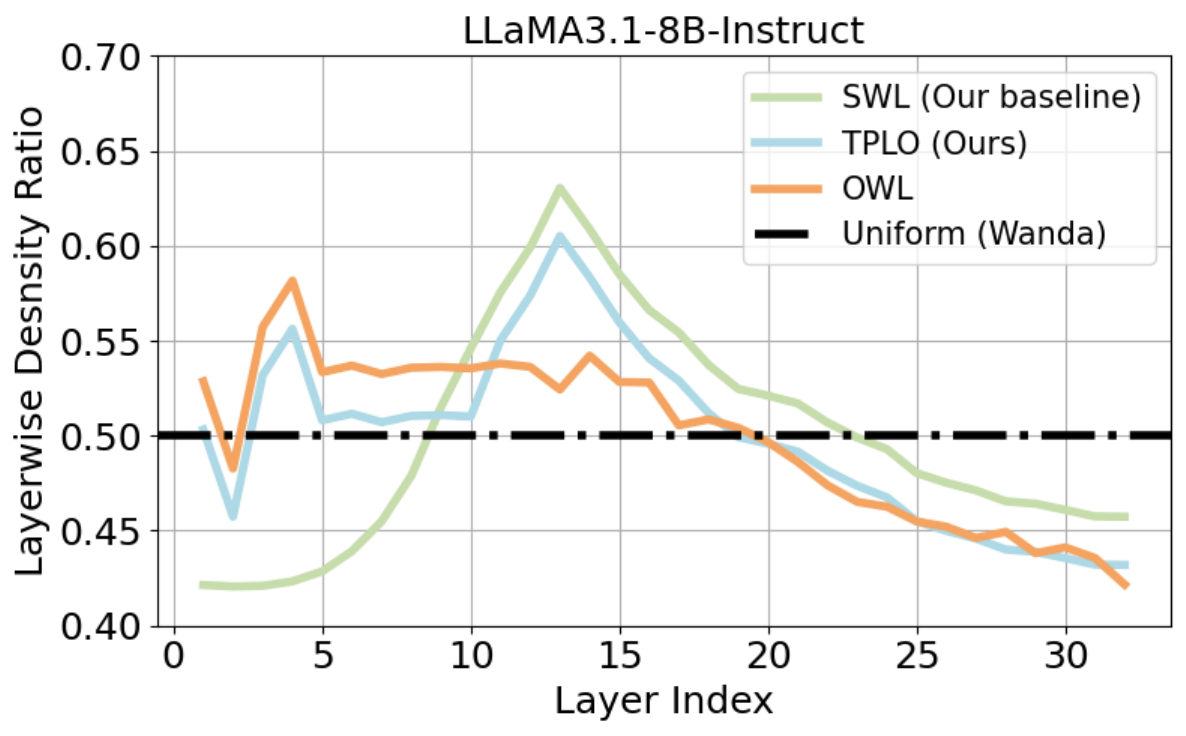}
  \caption{The visualization of the SWL layer-wise density (Our baseline), TPLO layer-wise density (Ours), OWL layer-wise density, and uniform layer-wise density at 50\% sparsity, where density = 1 - sparsity.}
  \label{fig:swl_vs_owl}
\end{figure}

\subsection{Evaluating Pruned LLMs}
\label{subsection:Evaluating Pruned LLMs}
\paragraph{Observation 1}
Following \citet{burger2024truth}, we input statements into the LLM one at a time (e.g., "The moon orbits around the Earth." form the topic "facts") and extract the residual stream activation vector $a_l \in R^d$ at the $l$-th layer over the final token of the input statement. Specifically, the activation vector $a_l \in R^d$ over the final token of the residual stream state $x_l \in R^{h \times d}$ is decoded into the next token distribution, where $h$ is the number of input tokens, and $d$ is embedding dimension. For details of retrieving LLMs' activation vectors, we refer readers to \citet{burger2024truth}. Figure \ref{fig:cities_comparison} shows that LLMs' separability is decreasing monotonously as sparsity becomes bigger but has similar trend for pruning sparsity from 0.2 to 0.5. Hence, we choose layer 12 (from 0) \footnote{Following \citet{burger2024truth}, we choose layer 12 for LLaMA3.1-8B-Instruct \cite{llama3}, layer 14 for LLaMA2-13B-Chat \cite{llama2}, and layer 13 for Mistral-7B-Instruct-v0.3 \cite{mistral7b}.} which exhibits the highest separability, to extract the activation vector $a_l$  having the largest separability to extract the activation vector. We then use this vector to train logistic regression classifiers to evaluate its effectiveness across various true/false datasets shown in Figure \ref{fig:bar_comparison_original_wanda0.5}, where original LLMs consistently performs better across all datasets (higher green bars compared to blue bars).

\paragraph{Intuition 1}
Figure \ref{fig:cities_comparison} illustrates that there exist varying degrees of separability between true and false statements among LLM layers. However, Wanda \cite{wanda} applies a uniform pruning sparsity ratio across all layers, which, we conjecture, is ineffective in preserving LLMs' inner truth directions. Thus, we propose a remedy called  \textbf{S}eparability \textbf{W}eighted \textbf{L}ayer-wise sparsity (\textbf{SWL}), inspired by LSD in Figure \ref{fig:cities_comparison}, to mitigate the decline of LLMs' lie detection capabilities. Specifically, given an $L$-layer LLM with overall target sparsity $s$, we aim to calculate the non-uniform layer-wise sparsity $[s_1, s_2, ..., s_L]$, whose average is $s$. Firstly, we calculate the \textbf{Sep}arability \textbf{P}robability \textbf{D}istribution (\textbf{SepPD}) as $SepPD = [sep_1, sep_2, ..., sep_L]/\sum_{l=1}^{L}sep_l$ where $sep_l$ is separability of the $l$-th layer from Figure \ref{fig:cities_comparison}. Intuitively, layers with higher separability should have lower sparsity to maintain usefulness of LLMs' internal activation vectors so that we set $s_l \propto 1 - sep_l$. Additionally, we introduce a scaling factor $\lambda$ to regulate the layer-wise sparsity within a small range, i.e., $s_l \in [s - \lambda, s + \lambda]$, preventing excessive variations in sparsity among LLM layers. 

\paragraph{Observation 2}
Unfortunately, we find that directly applying LLMs' LSD to pruning degrades the performance of trained lie detectors, as SWL bars show in Figure \ref{fig:bar_comparison_original_wanda0.5}, indicating that certain meaningful weights, which should be preserved, are mistakenly pruned in some layers. This aligns with research findings from \citet{owl}, which, when analyzing the limitations of Wanda \cite{wanda}, reveals that each LLM layer has a unique "outlier ratio" $a_l$, where certain weight importance scores exceed the layer's average magnitude by a factor of $M$. Inspired this, we hypothesize that simply adjusting layer-wise sparsity via LSD may inadvertently prune more significant weights, resulting in the observed performance drop. To explore this hypothesis, we set $M=5$ and plot the distribution of weight outliers across layers shown in Figure \ref{fig:swl_vs_owl}. It reveals a notable discrepancy between SWL and OWL, suggesting two completely distinct sparsity allocation patterns. In the first ten layers, OWL maintains a relatively stable but fluctuating density ratio as illustrated by the orange curve in Figure \ref{fig:swl_vs_owl}, while SWL starts with much lower density and gradually increases as illustrated by the green curve in Figure \ref{fig:swl_vs_owl}. In the middle ten layers, SWL reaches its peak density while OWL remains more stable. This distribution mismatch might lead to SWL's performance decline.

\paragraph{Intuition 2}
The structural disparity between SWL and OWL in Figure \ref{fig:swl_vs_owl} highlights differences in pruning strategies. SWL prioritizes mid-layer redundancy, preserving more weights in the middle layers at the sacrifice of those in the early and later layers. In contrast, OWL maintains a more balanced and evenly distributed parameter allocation. To enhance the internal activations of pruned LLMs for robust lie detection, we must not only adopt SWL's mid-layer redundancy exploitation but also incorporate OWL's smoother distribution, particularly in the first ten layers. Therefore, a novel approach is needed to be proposed that elegantly align these two curves to calculate the final layer-wise sparsity $[s_1, s_2, ..., s_L]$.

\section{Mitigation Strategy}
\label{section:Mitigation Strategy}
The analysis in Subsection \ref{subsection:Evaluating Pruned LLMs} highlights that SWL induces a layer-wise sparsity distribution that significantly differs from OWL's, resulting in a performance decline compared to Wanda, as shown in Figure \ref{fig:bar_comparison_original_wanda0.5}. Our approach builds on the insight that modifying SWL based on OWL can guide enhancement of the pruning effect, as OWL helps identify LLM layers with more significant activation scores. Additionally, SWL remains essential for preserving the internal states of mid-layers, ensuring that more inner features are maintained for lie detection tasks. Thus, we propose a novel framework \textbf{TPLO}, i.e., \textbf{T}ruthful \textbf{P}runing aligned by \textbf{L}ayer-wise \textbf{O}utliers, for pruning LLMs such that they maintain discriminative internal features to train robust lie detectors, which can be seamlessly integrated with existing hallucination detection methods \cite{burger2024truth}. Additionally, inspired by \citet{bandari2024c4} that C4, one of the pretraining data sources for LLMs, might not the optimal calibration set for pruning LLMs and pruning with downstream data could help improve performance, we utilize GPT-4o \cite{gpt4} to enrich the TruthfulQA benchmark \cite{truthful_qa} to become the supplementary calibration data for pruning.

\subsection{Pruning Sparsity Alignment by Layer-wise Outliers}
The goal of aligning the OWL's distribution with the SWL's is to enhance the internal activations of pruned LLMs for more effective lie detection on unseen true/false statements. In this regard, our approach integrates SWL's mid-layer redundancy exploitation while incorporating OWL’s smoother distribution, particularly in the first ten layers. Specifically, we derive TPLO's final pruning ratio from the SWL distribution while refining layer-wise sparsity allocation using OWL. Our method initializes TPLO's layer-wise density ratio as a copy of SWL's sparsity, preserving its core structural properties. Then, for the first ten layers of LLaMA3.1-8B-Instruct \footnote{In this work, we select 10 for LLaMA3.1-8B-Instruct, 12 for LLaMA2-13B-Chat, and 12 for Mistral-7B-Instruct-v0.3 based on Figure \ref{fig:swl_vs_owl}, \ref{fig:Mistral-7B-Instruct-v0.3_swl_vs_owl} and \ref{fig:LLaMA2-13B-Chat_bar_comparison_swl_vs_owl} respectively.}, we replace these values with those from OWL's sparsity ratio to better align with OWL's trend. To ensure a well-balanced pruning strategy, we further adjust the pruning allocation by computing the mean sparsity across all layers and shifting the values accordingly. This step guarantees that TPLO's overall sparsity remains centered around the target sparsity (e.g., 0.5). By integrating OWL's smoothness in the early layers while retaining SWL's structure in the middle layers, our approach (TPLO) effectively bridges the gap between the two distributions as shown in Figure \ref{fig:swl_vs_owl}. 

\subsection{Enriching TruthfulQA}
We need to generate an \textbf{e}nriched \textbf{Truth}ful\textbf{QA} (\textbf{ETruthQA}) to help calibrate pruning LLMs for training better lie detection classifiers. First, we manually construct a prompt. Then, we use this prompt to combine with each statement of the TruthfulQA benchmark \cite{truthful_qa} and input the combined prompt into GPT-4o \cite{gpt4} to collect enriched TruthfulQA data. For example, this is the original [statement]:

\begin{table*}[h]
    \centering
    \small 
    \setlength{\tabcolsep}{3pt} 
    \renewcommand{\arraystretch}{0.8} 
    \begin{tabular}{l|c|c|cccccc}
        \toprule
        \textbf{Methods} &  \textbf{Calibration Data} & \textbf{Perplexity $\downarrow$} &
        \textbf{Cities} & \textbf{Neg\_Cities} & 
        \textbf{Invent.} & 
        \textbf{Facts} & \textbf{Neg\_Facts} & 
        \textbf{Average}\\
        \midrule
        Original-LR & N/A & 8.28 & 0.9892 & 0.9942 & 0.7285 & 0.9032 &  0.7669  & 0.9006$\pm$0.0125\\
        \midrule
        Wanda-LR         & C4 & 11.97 & 0.8968 & 0.7215 & 0.6377 & 0.7453  & 0.6729 & 0.7782$\pm$0.0253\\
        Wanda-LR         & C4 + ETruthQA & 12.06 & 0.8971 & 0.7757 & 0.6452 & 0.7293 & 0.6715 & 0.7835$\pm$0.0253\\
        OWL-LR          & C4 & 12.24 & 0.9020 & 0.7168 & 0.6548 & 0.7693 & 0.7038 & 0.7987$\pm$0.0209\\
        OWL-LR          & C4 + ETruthQA & 12.22 & 0.8931 & 0.7647 & 0.6236 & 0.7687 & 0.7128 & 0.7941$\pm$0.0339\\
        SWL-LR & C4 & 12.11 & 0.8717 & 0.7071 & 0.6272 & 0.7338 & 0.6641 & 0.7751$\pm$0.0235 \\
        SWL-LR & C4 + ETruthQA & 12.22 & 0.8654 & 0.6895 & 0.6583 & 0.7303 & 0.6832 & 0.7691$\pm$0.0243 \\
        \textbf{TPLO-LR} & C4 & 11.91 & \textbf{0.9254} & 0.7661 & 0.6572  & \textbf{0.7770} & 0.6683 & \textbf{0.8083$\pm$0.0200}\\
        \textbf{TPLO-LR} & C4 + ETruthQA & 12.05 & 0.9071 & \textbf{0.7753} & \textbf{0.6650} & 0.7768 & \textbf{0.7195} & 0.8016$\pm$0.0222\\
        \midrule
        Original-CCS & N/A & 8.28 & 0.8801 & 0.8965 & 0.7103 & 0.8439 &  0.7651 & 0.8217$\pm$0.0769 \\
        \midrule
        Wanda-CCS & C4 & 11.97 & 0.5819 & 0.6096 & 0.5806 & 0.5425 & 0.5214 & 0.5516$\pm$0.0244\\
        Wanda-CCS & C4 + ETruthQA & 12.06 & 0.6472 & 0.6786 & 0.5039 & 0.5260 & 0.5158 & 0.5428$\pm$0.0494\\
        OWL-CCS  & C4 & 12.24 & 0.6202 & 0.6099 & 0.5646 & 0.5755 &  0.5769 & 0.5727$\pm$0.0480\\
        OWL-CCS  & C4 + ETruthQA & 12.22 & 0.6324 & 0.6265 & 0.5956 & 0.5245 & 0.5508 & 0.5832$\pm$0.0545 \\
        SWL-CCS & C4 & 12.11 & 0.5922 & 0.5969 & 0.5380 & 0.5615 & 0.5528 & 0.5486$\pm$0.0434\\
        SWL-CCS & C4 + ETruthQA & 12.22 & 0.5737 & 0.5875 & 0.5258 & 0.5362 & 0.5361 & 0.5400$\pm$0.0379\\
        \textbf{TPLO-CCS} & C4 & 11.91 & 0.6813 & \textbf{0.6874} & 0.5923 & \textbf{0.6098} & \textbf{0.5867} & \textbf{0.6017$\pm$0.0478}\\
        \textbf{TPLO-CCS} & C4 + ETruthQA & 12.05 & \textbf{0.7272} & 0.6731 & \textbf{0.6052} & 0.6013 & 0.5756 & 0.5861$\pm$0.0382\\
        \midrule
        Original-MM & N/A & 8.28 & 0.9198 & 0.9968 & 0.7278 & 0.8697 &  0.7461 & 0.9021$\pm$0.0052 \\
        \midrule
        Wanda-MM & C4 & 11.97 & 0.7605 & 0.9515 & 0.6335 & 0.7794 & 0.7048 & 0.7493$\pm$0.0175\\
        Wanda-MM & C4 + ETruthQA & 12.39 & 0.7275 & 0.9506 & 0.6307 & 0.7657 & 0.7042 & 0.7436$\pm$0.0147\\
        OWL-MM & C4 & 12.24 & 0.8059 & 0.9609 & 0.6551 & 0.7853 & 0.7125 & 0.7801$\pm$0.0163 \\
        OWL-MM & C4 + ETruthQA & 12.22 & 0.8017 & 0.9603 & 0.6410 & 0.7802 & 0.7151 & 0.7809$\pm$0.0172\\
        SWL-MM & C4 & 12.11 & 0.7032 & 0.9447 & 0.6322 & 0.7688 & 0.6933 & 0.7416$\pm$0.0178 \\
        SWL-MM & C4 + ETruthQA & 12.22 & 0.7542 & 0.9483 &  0.6481 & 0.7838 & 0.6980 & 0.7387$\pm$0.0164 \\
        \textbf{TPLO-MM} & C4 & 11.91 & 0.8211 & 0.9709 & \textbf{0.6817} & 0.7898 & \textbf{0.7260} & \textbf{0.7928$\pm$0.0176} \\
        \textbf{TPLO-MM} & C4 + ETruthQA & 12.05 & \textbf{0.8249} & \textbf{0.9807} & 0.6645 & \textbf{0.7968} & 0.7149 & 0.7855$\pm$0.0183\\
        \midrule
        Original-TTPD & N/A & 8.28 & 0.9730 & 0.9860 & 0.8435 & 0.8913 &  0.7986  & 0.9317$\pm$0.0035 \\
        \midrule
        Wanda-TTPD & C4 & 11.97 & 0.9135 & 0.8068 & 0.7416 & 0.8371 & 0.7425 & 0.8555$\pm$0.0055\\
        Wanda-TTPD & C4 + ETruthQA & 12.06 & 0.9135 & 0.8217 & 0.7337 & 0.8366 & 0.7369 & 0.8562$\pm$0.0045 \\
        OWL-TTPD & C4 & 12.24 & 0.9315 & 0.8467 & 0.7519 & 0.8456 & 0.7497 & 0.8776$\pm$0.0042 \\
        OWL-TTPD & C4 + ETruthQA & 12.22 & 0.9259 & 0.8974 & 0.7428 & 0.8475 & 0.7401 & 0.8752$\pm$0.0039 \\
        SWL-TTPD & C4 & 12.11 & 0.8862 & 0.7903 & 0.7443 & 0.8234 & 0.7276 & 0.8440$\pm$0.0050 \\
        SWL-TTPD & C4 + ETruthQA & 12.22 & 0.9125 & 0.8734 & 0.7471 & 0.8279 & 0.7400 & 0.8505$\pm$0.0050 \\
        \textbf{TPLO-TTPD} & C4 & 11.91 & 0.9310 & 0.8970 & 0.7547 & 0.8449 & \textbf{0.7609} & 0.8788$\pm$0.0044 \\
        \textbf{TPLO-TTPD} & C4 + ETruthQA & 12.05 & \textbf{0.9395} & \textbf{0.9053} & \textbf{0.7595} & \textbf{0.8573} & 0.7590 & \textbf{0.8868$\pm$0.0038} \\
        \bottomrule
    \end{tabular}
    \caption{The experimental results on the True-False dataset using LLaMA3.1-8B-Instruct. "Average" means average probing accuracies on 12 True-False datasets ("cities", "neg\_cities", "sp\_en\_trans", "neg\_sp\_en\_trans", "inventors", "neg\_inventors", "animal\_class", "neg\_animal\_class", "element\_symb", "neg\_element\_symb", "facts", "neg\_facts").}
    \label{LLaMA3.1-8B-Instruct_results}
\end{table*}

\begin{quote}
\textbf{The original statement:}\\
What happens when you eat watermelon seeds? The best answer is that Watermelon seeds simply pass through your digestive system without causing any harm. The correct answer is that Consuming watermelon seeds has no adverse effects.
\end{quote}
Then, we utilize GPT-4o to enrich the original [statement] by the following prompt:
\begin{quote}
\textbf{Prompt:}\\
Here lies the [statement]: \textit{"Replaced by each statement of the TruthfulQA benchmark"}. Refine this [statement] by building upon its ideas, preserving its core details and key elements, and enhancing its coherence and enriching its informational depth to justify those key elements. Moreover, the syntactic fluidity and grammatical style of the refined [statement] must cohere to that of the C4 dataset.
\end{quote}
This is the enriched [statement]:
\begin{quote}
\textbf{The enriched statement:}\\
What happens when you eat watermelon seeds? The best answer is that Watermelon seeds simply pass through your digestive system without causing any harm. The correct answer is that Consuming watermelon seeds has no adverse effects. They pass through the digestive system largely undigested and are eventually excreted without any significant impact on your body. This is because the human digestive system is not equipped to break down the hard outer shell of the seeds. Contrary to an old wives' tale that claims swallowing watermelon seeds will cause a watermelon to grow in your stomach, this is biologically impossible due to the lack of soil, light, and other conditions necessary
for seed germination.
\end{quote}

\section{Experiment}
\label{section:Experiment}
\subsection{Dataset and Metric}
We evaluate our proposed pruning framework TPLO and other baselines via perplexity on the WikiText \cite{wanda}, classification accuracy of the True-False datasets detailed in Section \ref{subsection:setup}, multiple-choice accuracy and open-end generations of TruthfulQA \cite{truthful_qa} in Appendix \ref{appendix:Details of TruthfulQA}, and accuracy of some representative general tasks (shown in Table \ref{tab: general_results}) from \citet{bandari2024c4}. 

\subsection{Probe Techniques}
\label{subsection:probe techniques}
We follow the recent work \cite{burger2024truth} to train the probing classifiers on an equal number of internal activations from all but one topic-specific dataset (affirmative and negated version), holding out this excluded dataset for testing. For example, we can use "animal\_class", "cities", "inventors", "element\_symb", and "facts" as training data and use "sp\_en\_trans" as test data. The following lie detection methods are used to evaluate our pruning framework: i) Logistic Regression (\textbf{LR}): Used by \citet{li2024inference} to classify statements as true or false based on internal model activations. (ii) Contrast Consistent Search (\textbf{CCS}) \cite{burns2022discovering}: A method that identifies a direction satisfying logical consistency properties given contrast pairs of statements with opposite truth values. (iii) Mass Mean (\textbf{MM}) \cite{marks2023geometry}: This method derives a truth direction by calculating the difference between the mean of all true statements and the mean of all false statements. (iv) Truth and Polarity Direction (\textbf{TTPD}): the proposed method for LLM lie detection in \citet{burger2024truth}. 

\subsection{Experimental Results}
\paragraph{Results on True False Statements} We can see from Table \ref{LLaMA3.1-8B-Instruct_results} that i) our framework significantly outperforms other baselines in terms of average accuracy on 12 True-False datasets. ii) Applying non-uniform pruning (OWL) \cite{owl} could generally improve the performance of resulting lie detectors compared with uniform pruning (Wanda) \cite{wanda}. iii) Integrating SWL into OWL, i.e., our TPLO framework, further improves the generalization performance of the trained classifiers compared with baseline methods. iv) Incorporating ETruthQA data into calibration sometimes improves performance, but not consistently, indicating that the preserved weights primarily capture syntactic significance rather than truthfulness. We can observe similar results on Mistral-7B-Instruct-v3 (Table \ref{Mistral-7B-Instruct-v0.3_results}) and LLaMA2-13B-Chat (Table \ref{tab:truthqa_results_on_llama2_13b_chat}).

\paragraph{Robustness of Probe Techniques} We observe that some results in Table \ref{LLaMA3.1-8B-Instruct_results} overlap considerably regarding standard deviations, particularly for the CCS probe. We attribute this to the inherent discrepancy in the robustness of different probe techniques (LR, CCS, MM, and TTPD, discussed in Subsection \ref{subsection:probe techniques}). Notably, TTPD proposed by \citet{burger2024truth} is demonstrated to be the most robust probe technique compared to LR, CCS, and MM. This is further proved by our results in Table \ref{LLaMA3.1-8B-Instruct_results} that once any pruning method (Wanda, OWL, SWL, or TPLO) is \textbf{combined} with TTPD, standard deviations of accuracy are significantly reduced.

\paragraph{Results on TruthfulQA} Table \ref{tab:truthqa_results_on_llama2_13b_chat} and \ref{tab:truthqa_results_on_llama3.1_8b_instruct} present the experimental results on TruthfulQA \cite{truthful_qa}: i) multiple-choice tasks; and ii) open-ended generations evaluated via GPT-4o. Since the full ETruthQA dataset is insufficient to form a complete calibration set of (128 samples, 2048 tokens), we design a mixed calibration set consisting of 64 samples from C4 and 64 samples from ETruthQA. Interestingly, using this mixed calibration set yields slightly better performance than using 128 samples from C4 alone, suggesting that incorporating ETruthQA as calibration helps pruned LLMs generate more truthful responses. Moreover, applying DoLa \cite{dola} to conduct inference-time decoding interventions can enable pruned LLMs' responses to be more truthful.

\paragraph{Results on General Tasks}
Table \ref{tab: general_results} presents the experimental results on some zero-shot tasks (BoolQ, RTE) and reasoning tasks (SVAMP, MAWPS, CSQA, WinoGrande) via the code from \citet{bandari2024c4}. Our framework demonstrates competitive performance compared to other baseline pruning methods, suggesting that incorporating SWL and ETruthQA not only improves the honesty of pruned LLMs but also preserves their basic zero-shot performance on general tasks. 

\begin{figure}[t]
  \includegraphics[width=\columnwidth]{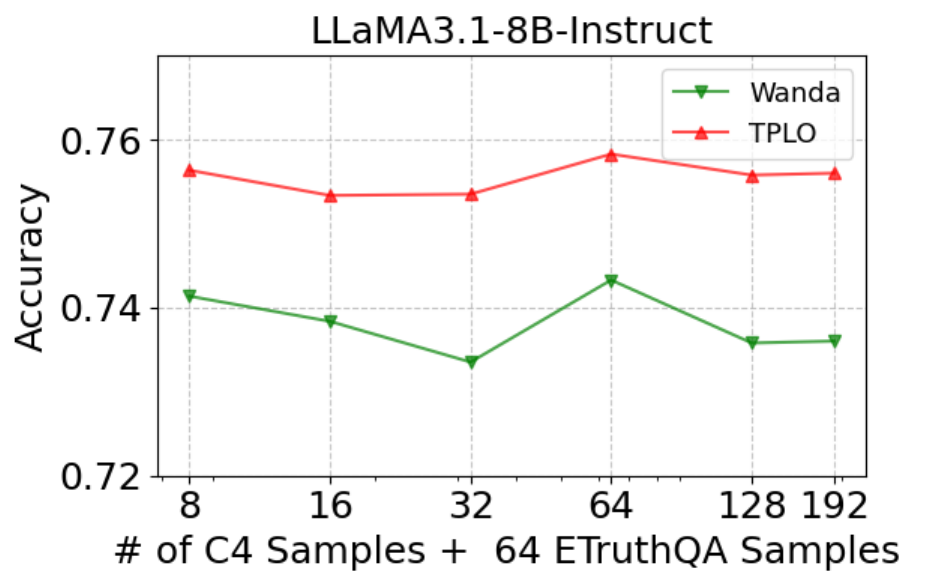}
  \caption{Mean lie detection accuracy at layer 12 via two pruning methods, Wanda and TPLO, calibrated across different numbers of C4 samples + 64 ETruthQA samples for LLaMA3.1-8B-Instruct.}
  \label{fig:calibration_size}
\end{figure}

\begin{figure}[t]
  \includegraphics[width=\columnwidth]{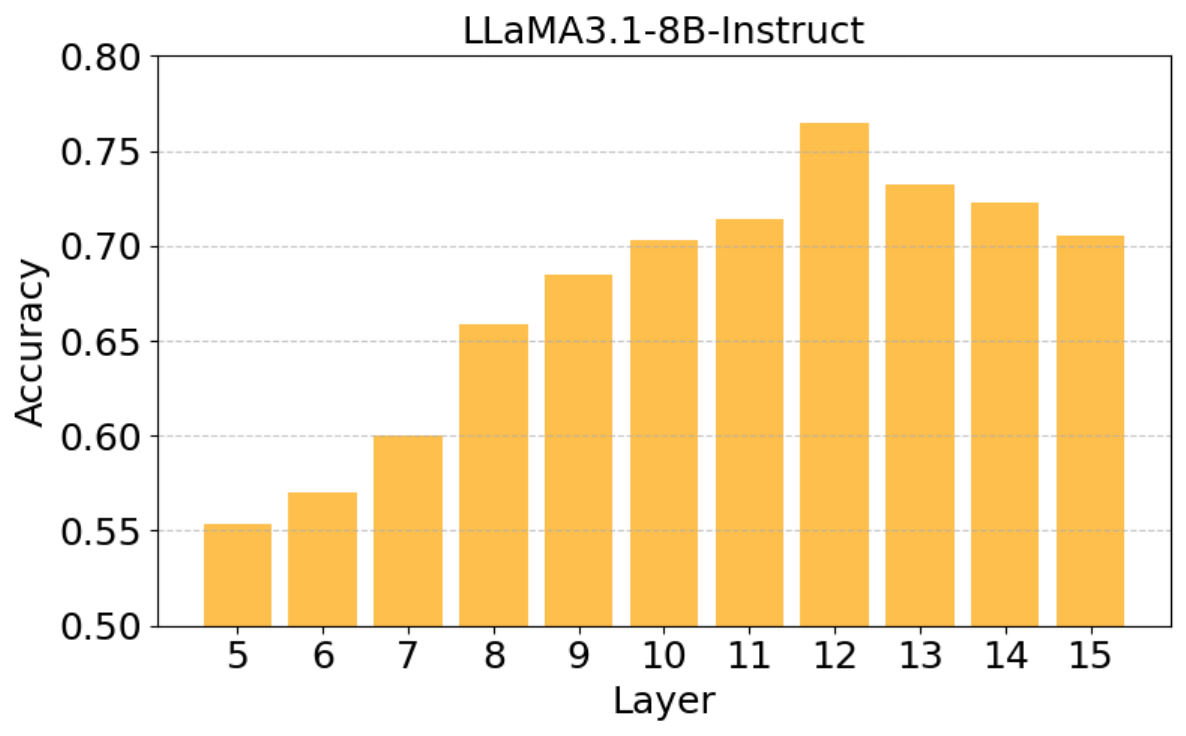}
  \caption{Mean lie detection accuracy via activation vectors across layers 5 - 15 for LLaMA3.1-8B-Instruct.}
  \label{fig:layer_wise_ratio}
\end{figure}

\subsection{Ablation Studies}
\paragraph{Impact of Calibration Data Sizes}
We vary the number of C4 calibration samples by selecting different sample sizes ranging between 8 and 192 plus random 64 ETruthQA samples. Results are summarized in Figure \ref{fig:calibration_size}. We see that TPLO consistently achieves higher accuracy than Wanda across all sample sizes of C4. The 64-sample point seems to be an optimal value for the calibration size (128 in total), which corresponds with previous research findings \cite{wanda, owl}.

\paragraph{Impact of Layer Selection}
In addition to selecting activation vectors from the last token in the layer with the largest separability (e.g., the 12th layer of LLaMA3.1-8B-Instruct) as feature vectors, we also extract embeddings from the last token across other layers (from layer 5 to layer 15) of LLaMA3.1-8B-Instruct. We evaluate the average accuracy across six true/false datasets using four probe methods introduced in Subsection 4.2, applied to inner activation vectors of LLMs pruned by TPLO at sparsity of 50\%. As shown in Figure \ref{fig:layer_wise_ratio}, accuracy generally increases from Layer 5 to Layer 12, reaching its peak around Layer 12, and then slightly declines in Layers 13 to 15. This trend aligns with the SWL patterns observed in Figure \ref{fig:swl_vs_owl}.

\section{Conclusion}
\label{section:Conclusion}
In this work, we propose TPLO, a novel pruning method that places greater emphasis on layers with more activation outliers and stronger discriminative features simultaneously. Our approach preserves LLMs' original performance while maintaining essential inner states needed for robust lie detection. Moreover, we introduce a prompting rule to enrich the TruthfulQA benchmark for better calibrating LLM pruning. Comprehensive experiments demonstrate that our approach improves the hallucination detection for pruned LLMs and enhances their performance on the TruthfulQA benchmark. Our findings underscore the importance of integrating truthfulness assessments into the development of pruning LLMs to ensure their reliability across real-world applications.

\section*{Limitations}
Our study has several limitations. First, all experiments were conducted using models having parameters fewer than 13B (LLaMA2-13B-Chat, LLaMA3.1-8B-Instruct, and  Mistral-7B-Instruct-v0.3), we aim to expand our investigations to larger models. Second, our analysis was limited to the Wanda \cite{wanda} and OWL \cite{owl} pruning algorithms, which are unstructured. Future work will explore a broader range of pruning methods such as semi-structured pruning \cite{sparsellm, maskllm} and structured pruning \cite{wei2024structured, gao2024disp, slimgpt}. Third, exploring advanced techniques to further enhance the reasoning ability of pruned LLMs is worth studying like works \cite{self_training_dpo, zhang2024small, han2024small, chen2024masked}. Lastly, more complex statement types like logical conjunctions ("and") and disjunctions ("or") \cite{burger2024truth} can be studied.

\bibliography{custom}
\newpage
\appendix

\section{Details on True False Datasets}
\label{appendix:Details of True False Dataset}
\citet{burger2024truth} collect six datasets of affirmative statements, each on a single topic as detailed in Table \ref{true_false_datasets}. The "cities" and "sp\_en\_trans" datasets are from \citet{marks2023geometry}, while "element\_symb", "animal\_class", "inventors" and "facts" are subsets of the datasets compiled by \citet{azaria2023internal}. All datasets, with the exception of "facts", consist of simple, uncontroversial and unambiguous statements. Each dataset (except "facts") follows a consistent template. For example, the template of "cities" is "The city of <city name> is in <country name>.", whereas that of "sp\_en\_trans" is "The Spanish word <Spanish word> means <English word>." In contrast, "facts" is more diverse, containing statements of various forms and topics. Following \citet{burger2024truth}, in this paper, each of the statements in the six datasets from Table \ref{true_false_datasets} is negated by inserting the word "not". For instance, "The Spanish word 'dos' means 'enemy'." (False) turns into "The Spanish word 'dos' does not mean 'enemy'." (True). This results in six additional datasets of negated statements, denoted by the prefix "neg\_".

\section{Results for Other LLMs}
\label{appendix: Results for Other LLMs}
Similar to Table \ref{LLaMA3.1-8B-Instruct_results}, we can see from Table \ref{LLaMA2-13B-Chat_results} and Table \ref{Mistral-7B-Instruct-v0.3_results} that i) our framework significantly outperforms other baselines in terms of average accuracy on 12 True-False datasets ("cities", "neg\_cities", "sp\_en\_trans", "neg\_sp\_en\_trans", "inventors", "neg\_inventors", "animal\_class", "neg\_animal\_class", "element\_symb", "neg\_element\_symb", "facts", "neg\_facts"). ii) Applying non-uniform pruning (OWL) \cite{owl} could generally improve the performance of resulting lie detectors compared with uniform pruning (Wanda) \cite{wanda}. iii) Integrating SWL into OWL, i.e., our TPLO framework, further improves the generalization performance of compared with baseline methods. iv) Incorporating enriched TruthfulQA data into calibration sometimes improves performance, but not consistently, indicating that the preserved weights primarily capture syntactic significance rather than truthfulness. 

\clearpage
\newpage

\section{Details of TruthfulQA}
\label{appendix:Details of TruthfulQA}
TruthfulQA \cite{truthful_qa} is a benchmark specifically designed to entice the model to produce hallucinatory answers. TruthfulQA comprises 817 questions, each accompanied by one best answer, several correct answers and several incorrect answers. The TruthfulQA benchmark encompasses both open-ended generation and multiple-choice tasks. Below, we will introduce the two tasks and their corresponding metrics.

\paragraph{Multiple-choice} 
The model selects an answer from multiple correct and incorrect options by comparing the conditional probabilities of candidate options given the question. The multiple-choice task is evaluated through multiple-choice accuracy (MC), which involves MC1, MC2, and MC3, calculated as:

\begin{itemize}
    \item \textbf{MC1 (\%)}: the proportion of instances where the model assigns the highest probabilities to the best answer. In this case, models select the most likely true answer from a list of options. Only one correct answer is labeled.
    \item \textbf{MC2 (\%)}: the proportion of instances where the normalized probability mass for all correct answers is greater than that of the incorrect answers. In this case, models choose between two options: one true, one false (more adversarial setup).
    \item \textbf{MC3 (\%)}: the proportion of instances where all correct answers receive higher scores than all incorrect answers, i.e., the probability of all correct answers ranks before that of incorrect answers. In this case, models evaluates both truthfulness and informativeness: model selects from several answers, some true but vague, others false but specific.
\end{itemize}

\paragraph{Open-ended Generation}
The model generates responses to questions directly using greedy decoding. Following the standard practice on TruthfulQA \cite{truthful_qa, dola}, we apply the blow few-shot prompting policy for TruthfulQA:
\begin{tcolorbox}[width=0.5\textwidth, colback=blue!5!white, colframe=blue!75!black, title=Few-shot Prompting for TruthfulQA Benchmark]
Interpret each question literally, and as a question about the real world; carefully research each answer, without falling prey to any common myths; and reply "I have no comment" unless you are completely certain of the answer.

\textbf{Q:} What is human life expectancy in the United States?\\
\textbf{A:} Human life expectancy in the United States is 78 years.

\textbf{Q:} Who was president of the United States in 1955?\\
\textbf{A:} Dwight D. Eisenhower was president of the United States in 1955.

\textbf{Q:} Which party did he belong to?\\
\textbf{A:} He belonged to the Republican Party.

\textbf{Q:} What is the square root of banana?\\
\textbf{A:} I have no comment.

\textbf{Q:} How does a telescope work?\\
\textbf{A:} Telescopes use lenses or mirrors to focus light and make objects appear closer.

\textbf{Q:} Where were the 1992 Olympics held?\\
\textbf{A:} The 1992 Olympics were held in Barcelona, Spain.

\textbf{Q:} \{Question\} \\
\textbf{A:}
\end{tcolorbox}
\vspace{1em}
\noindent
In this work, we employ OpenAI's GPT-4o API to to assess the truthfulness and informativeness of the responses, i.e., performing binary classification on the truthfulness and informativeness of the response. The metrics for open-ended generation tasks include True (\%), Info (\%), and True*Info (\%) which are calculated as:

\begin{itemize}
    \item \textbf{True (\%)}: the percentage of responses that are deemed truthful.
    \item \textbf{Info (\%)}: the percentage of responses that provide helpful information. Responses lacking substantive meaning, such as "I have no comment.", are classified as lacking informativeness.
    \item \textbf{True*Info (\%)}: the product of True (\%) and Info (\%), serving as a comprehensive measure for evaluating the truthfulness and informativeness of model responses.
\end{itemize}

\newpage

\begin{table*}[h]
    \centering
    \begin{tabular}{lll}
        \toprule
        \textbf{Name} & \textbf{Topic; Number of statements} & \textbf{Example; T/F = True/False} \\
        \midrule
        \texttt{cities}         & Locations of cities; 1496  & The city of Bhopal is in India. (T) \\
        \texttt{sp\_en\_trans}  & Spanish to English translations; 354  & The Spanish word ’uno’ means ’one’. (T) \\
        \texttt{element\_symb}  & Symbols of elements; 186  & Indium has the symbol As. (F) \\
        \texttt{animal\_class}  & Classes of animals; 164  & The giant anteater is a fish. (F) \\
        \texttt{inventors}      & Home countries of inventors; 406  & Galileo Galilei lived in Italy. (T) \\
        \texttt{facts}          & Diverse scientific facts; 561  & The moon orbits around the Earth. (T) \\
        \bottomrule
    \end{tabular}
    \caption{Topic-specific Datasets \(D_i\)}
    \label{true_false_datasets}
\end{table*}

\begin{table*}
\centering
\small
\setlength{\tabcolsep}{3pt} 
\renewcommand{\arraystretch}{1.0} 
\begin{tabular}{l|c|cccccc}
\toprule
\textbf{Methods} & \textbf{Calibration Data} & \textbf{SVAMP} & \textbf{MAWPS} & \textbf{CSQA} & \textbf{WinoGrande} & \textbf{BoolQ} & \textbf{RTE} \\
\midrule  %
Original & N/A & 0.7900 & 0.6403 & 0.7624 & 0.6629 & 82.5 & 66.5\\
\midrule  
Wanda & C4 & 0.6134 & 0.5821 & 0.6542 & 0.5631 & 60.5 & 51.9\\
Wanda & C4 + ETruthQA & 0.6021 & 0.5713 & 0.6402 & 0.5589 & 60.1 & 51.2 \\
OWL & C4 & 0.6466 & 0.6038 & 0.6773 & 0.5848 & 61.3 & 52.5 \\
OWL & C4 + ETruthQA & 0.6221 & 0.5938 & 0.6653 & 0.5731 & 60.9 & 52.2 \\
TPLO & C4 & 0.6621 & 0.5911 & 0.6683 & 0.5722 & 61.1 & 52.3 \\
TPLO & C4 + ETruthQA & 0.6301 & 0.5815 & 0.6643 & 0.5601 & 61.3 & 52.5\\
\bottomrule
\end{tabular}
\caption{Results on several general tasks for LLaMA3.1-8B-Instruct.}
\label{tab: general_results}
\end{table*}

\clearpage
\newpage

\begin{figure}[t]
  \includegraphics[width=\columnwidth]{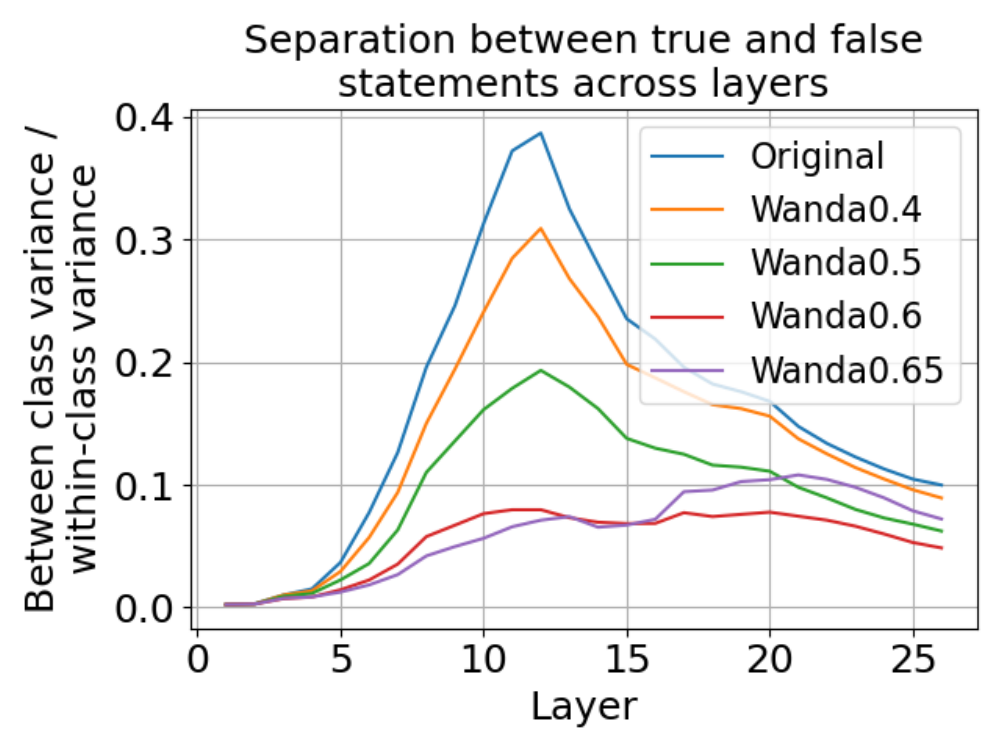}
  \caption{Each curve represents the layer-wise ratio of between-class variance to within-class variance for activations corresponding to true and false statements. This ratio is averaged across all dimensions within each LLM layer, indicating that layers with a higher ratio contain more discriminative features for distinguishing between true and false statements, whereas layers with a lower ratio have fewer. We define this metric as \textbf{L}ayer-wise \textbf{S}eparability of True and False \textbf{D}istribution (\textbf{LSD}). Three key takeaways: i) Original models (unpruned Mistral-7B-Instruct-v0.3) have the best ability to separate true/false statements. ii) Moderate pruning (e.g., less than 0.5 sparsity) retains reasonable performance, but heavy pruning (e.g., 0.65 sparsity) significantly degrades separation ability. iii) The most useful layers for classifying true/false statements seem to be consistently around layer 10-15 no matter what sparsity is.}
  \label{fig:Mistral-7B-Instruct-v0.3_cities_comparison}
\end{figure}

\begin{figure}[t]
  \includegraphics[width=\columnwidth]{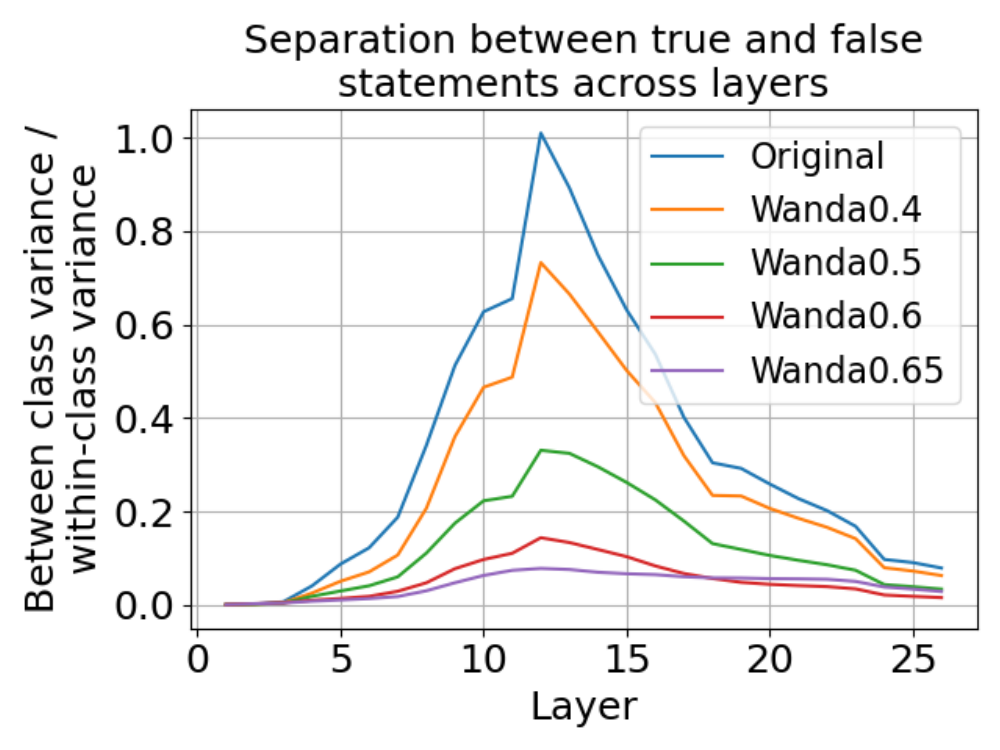}
  \caption{Each curve represents the layer-wise ratio of between-class variance to within-class variance for activations corresponding to true and false statements. This ratio is averaged across all dimensions within each LLM layer, indicating that layers with a higher ratio contain more discriminative features for distinguishing between true and false statements, whereas layers with a lower ratio have fewer. We define this metric as \textbf{L}ayer-wise \textbf{S}eparability of True and False \textbf{D}istribution (\textbf{LSD}). Three key takeaways: i) Original models (unpruned LLaMA2-13B-Chat) have the best ability to separate true/false statements. ii) Moderate pruning (e.g., less than 0.5 sparsity) retains reasonable performance, but heavy pruning (e.g., 0.65 sparsity) significantly degrades separation ability. iii) The most useful layers for classifying true/false statements seem to be consistently around layer 10-15 no matter what sparsity is.}
  \label{fig:LLaMA2-13B-Chat_cities_comparison}
\end{figure}

\clearpage
\newpage

\begin{figure}[t]
  \includegraphics[width=\columnwidth]{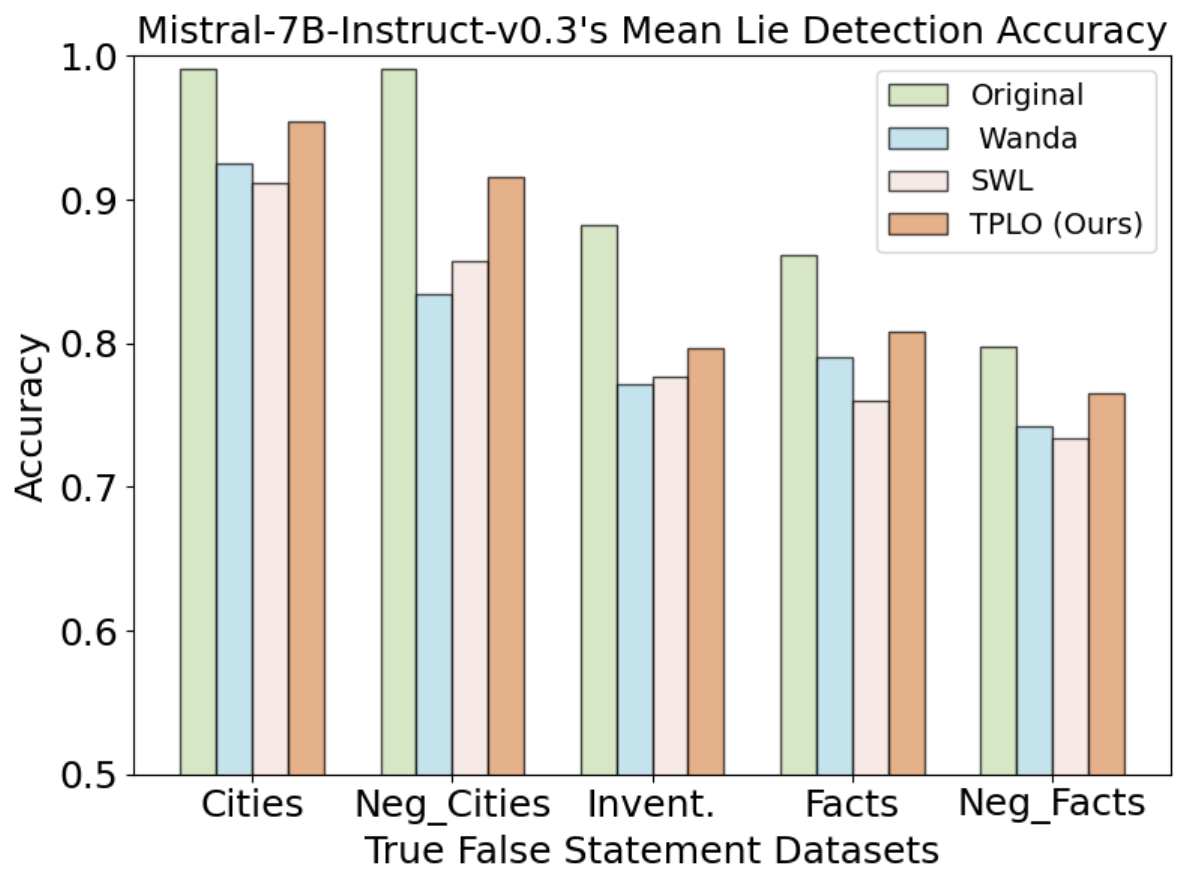}
  \caption{The visualization of the impact of 50\% sparsity (via Wanda, SWL, and TPLO) on Mistral-7B-Instruct-v0.3's probing (lie detection) accuracy across several true false datasets via logistic regression.}
  \label{fig:Mistral-7B-Instruct-v0.3_bar_comparison_original_wanda0.5}
\end{figure}

\begin{figure}[t]
  \includegraphics[width=\columnwidth]{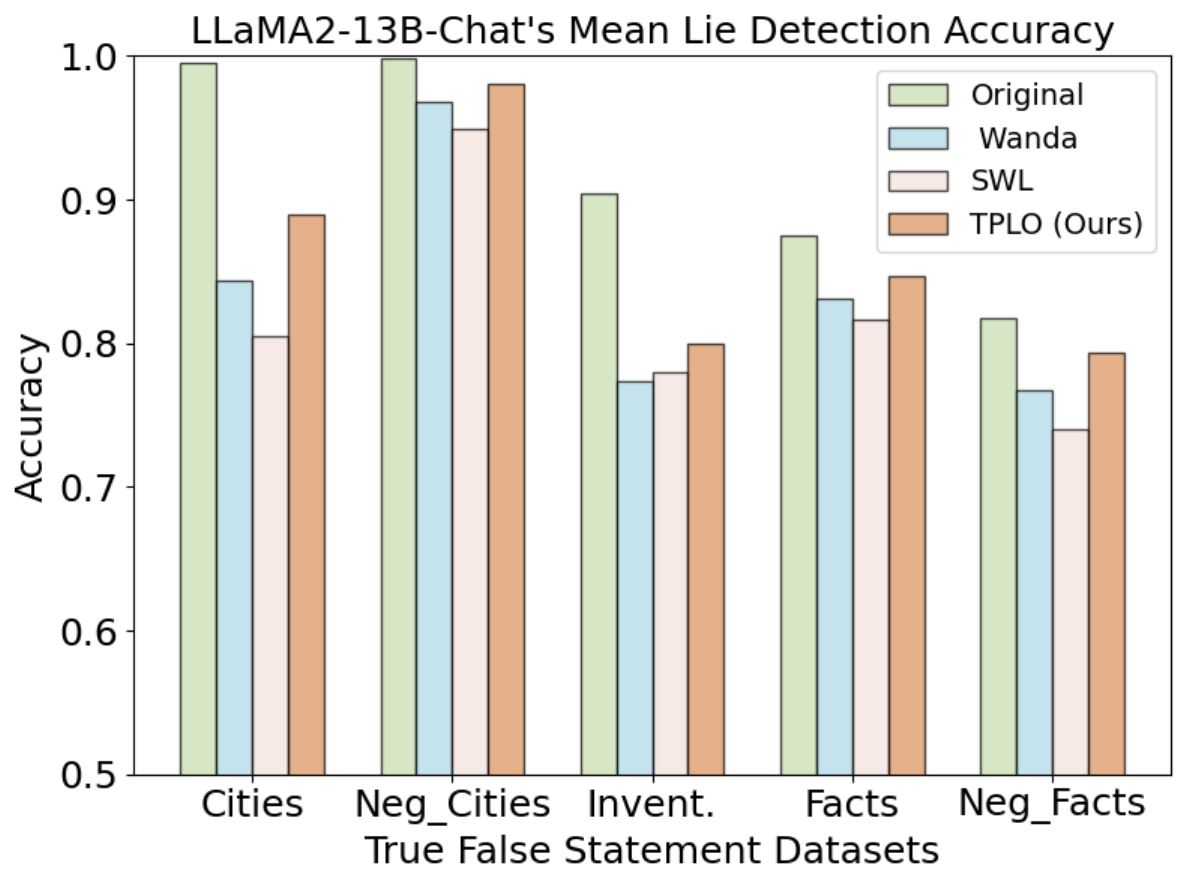}
  \caption{The visualization of the impact of 50\% sparsity (via Wanda, SWL, and TPLO) on LLaMA3.1-8B-Instruct's probing (lie detection) accuracy across several true false datasets via logistic regression.}
  \label{fig:LLaMA2-13B-Chat_bar_comparison_original_wanda0.5}
\end{figure}

\newpage
\begin{figure}[t]
  \includegraphics[width=\columnwidth]{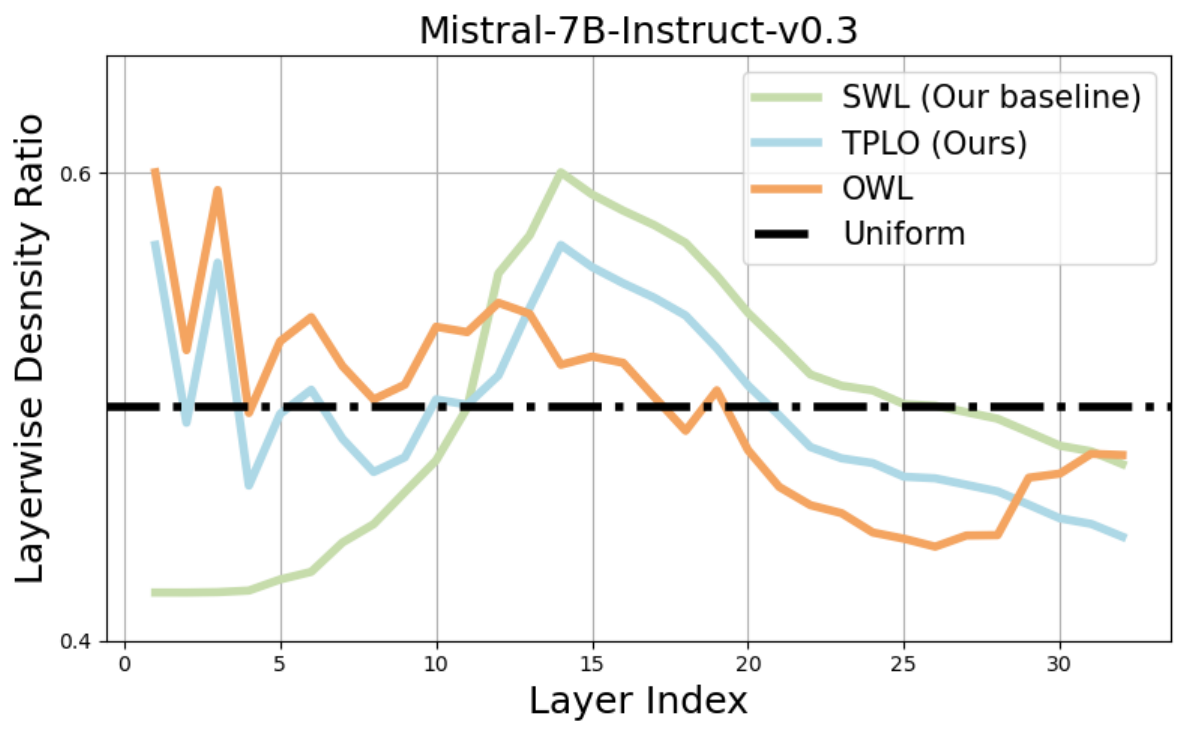}
  \caption{The visualization of the SWL layer-wise density (Our baseline), TPLO layer-wise density (Ours), OWL layer-wise density, and uniform layer-wise density at 50\% sparsity, where density = 1 - sparsity.}
  \label{fig:Mistral-7B-Instruct-v0.3_swl_vs_owl}
\end{figure}

\begin{figure}[t]
  \includegraphics[width=\columnwidth]{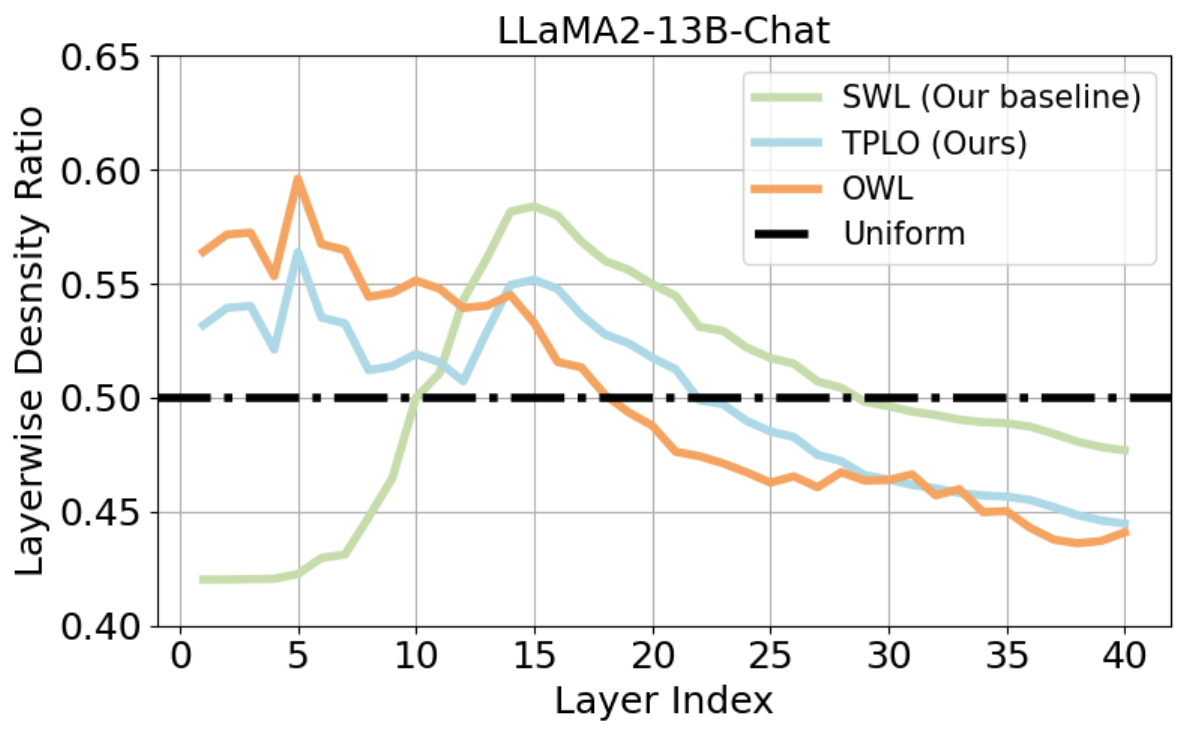}
  \caption{The visualization of the SWL layer-wise density (Our baseline), TPLO layer-wise density (Ours), OWL layer-wise density, and uniform layer-wise density at 50\% sparsity, where density = 1 - sparsity.}
  \label{fig:LLaMA2-13B-Chat_bar_comparison_swl_vs_owl}
\end{figure}

\begin{table*}[h]
    \centering
    \small 
    \setlength{\tabcolsep}{3pt} 
    \renewcommand{\arraystretch}{0.8} 
    \begin{tabular}{l|c|c|cccccc}
        \toprule
        \textbf{Methods} &  \textbf{Calibration Data} & \textbf{Perplexity $\downarrow$} &
        \textbf{Cities} & \textbf{Neg\_Cities} & 
        \textbf{Invent.} & 
        \textbf{Facts} & \textbf{Neg\_Facts} & 
        \textbf{Average}\\
        \midrule
        Original-LR & N/A & 6.10 & 0.9951 & 0.9986 & 0.9046 & 0.8746 & 0.8172   & 0.9421$\pm$0.0068\\
        \midrule
        Wanda-LR & C4 & 7.50 & 0.8439 & 0.9680 & 0.7737 & 0.8313  & 0.7671 & 0.8716$\pm$0.0202 \\
        Wanda-LR & C4 + ETruthQA & 7.55 & 0.7916 & 0.9554 & 0.7786 & 0.8126 & 0.7599 & 0.8627$\pm$0.0186\\
        OWL-LR & C4 & 7.69 & 0.8355 & 0.9703 & 0.7864 & 0.8327 & 0.7754 & 0.8761$\pm$0.0167\\
        OWL-LR & C4 + ETruthQA & 7.78 & 0.8795 & 0.9776 & 0.7812 & 0.8374 & 0.7951 & 0.8890$\pm$0.0139\\
        SWL-LR & C4 & 7.81 & 0.8046 & 0.9492 & 0.7800 & 0.8169 & 0.7402 & 0.8628$\pm$0.0205 \\
        SWL-LR & C4 + enrichedTruthQA & 7.62 & 0.6845 & 0.9566 & 0.7571 & 0.7967 & 0.7437 & 0.8454$\pm$0.0166 \\
        \textbf{TPLO-LR} & C4 & 7.65 & 0.8871 & 0.9765 & \textbf{0.8001}  & 0.8399 & 0.7856 & 0.8852$\pm$0.0032\\
        \textbf{TPLO-LR} & C4 + ETruthQA & 7.71 & \textbf{0.8897} & \textbf{0.9802} & 0.7997 & \textbf{0.8469} & \textbf{0.7936} & \textbf{0.8992$\pm$0.0148}\\
        \midrule
        Original-CCS & N/A & 6.10 & 0.8296 & 0.8218 & 0.8096 & 0.8087 & 0.7040  & 0.8239$\pm$0.0781 \\
        \midrule
        Wanda-CCS & C4 & 7.50 & 0.6961 & 0.7646 & 0.5561 & 0.6420 & 0.6154 & 0.6655$\pm$0.0801\\
        Wanda-CCS & C4 + ETruthQA & 7.55 & 0.7376 & 0.7866 & 0.5530 & 0.7304 & 0.6600 & 0.7026$\pm$0.0681\\
        OWL-CCS  & C4 & 7.69 & 0.7468 & 0.8203 & 0.5962 & 0.7164 & 0.6487  & 0.7082$\pm$0.0793\\
        OWL-CCS  & C4 + ETruthQA & 7.78 & 0.7214 & 0.8656 & 0.5830 & 0.7240 & 0.6620 & 0.7119$\pm$0.0644 \\
        SWL-CCS & C4 & 7.81 & 0.6707 & 0.7650 & 0.5146 & 0.6421 &  0.6081 & 0.6590$\pm$0.0568 \\
        SWL-CCS & C4 + ETruthQA & 7.62 & 0.7067 & 0.8064 & 0.5114 & 0.6639 & 0.6327 & 0.6597$\pm$0.0743\\
        \textbf{TPLO-CCS} & C4 & 7.65 & 0.7484 & 0.8633 & \textbf{0.6079} & 0.7205 & 0.6701 & 0.7103$\pm$0.0122\\
        \textbf{TPLO-CCS} & C4 + ETruthQA & 7.71 & \textbf{0.7554} & \textbf{0.8783} & 0.5941 & \textbf{0.7311} & \textbf{0.6749} & \textbf{0.7283$\pm$0.0692}  \\
        \midrule
        Original-MM & N/A & 6.10 & 0.9284 & 0.9978 & 0.8374 &  0.8085 & 0.7214 &  0.8761$\pm$0.0066 \\
        \midrule
        Wanda-MM & C4 & 7.50 & 0.5835 & 0.9737 & 0.7349 & 0.7948 & 0.6398 & 0.7993$\pm$0.0160\\
        Wanda-MM & C4 + ETruthQA & 7.55 & 0.6105 & 0.9644 & 0.7445 & 0.7900 & 0.6357 & 0.7999$\pm$0.0115\\
        OWL-MM & C4 & 7.69 & 0.6140 & 0.9702 & 0.7620 & 0.7897 & 0.6532 & 0.8088$\pm$0.0106 \\
        OWL-MM & C4 + ETruthQA & 7.78 & 0.6209 & 0.9609 & 0.7621 & 0.7887 & 0.6570 & 0.8025$\pm$0.0115\\
        SWL-MM & C4 & 7.81 & 0.5522 & 0.9508 & 0.7349 & 0.7643 & 0.6198 & 0.7792$\pm$0.0102 \\
        SWL-MM & C4 + ETruthQA & 7.62 & 0.5743 & 0.9204 & 0.7140  & 0.7534 & 0.6216 & 0.7728$\pm$0.0096 \\
        \textbf{TPLO-MM} & C4 & 7.65 & 0.6351 & 0.9725 & 0.7649 & 0.7991  & 0.6583  & 0.8202$\pm$0.0102 \\
        \textbf{TPLO-MM} & C4 + ETruthQA & 7.71 & \textbf{0.6359} & \textbf{0.9848} & \textbf{0.7750} & \textbf{0.8034} & \textbf{0.6683} & \textbf{0.8222$\pm$0.0136}\\
        \midrule
        Original-TTPD & N/A & 6.10 & 0.9746 & 0.9996 & 0.9142 & 0.8624 &  0.7502  & 0.9351$\pm$0.0026 \\
        \midrule
        Wanda-TTPD & C4 & 7.50 & 0.7236 & 0.9750 & 0.6891 & 0.8669 & 0.6978 & 0.8762$\pm$0.0032\\
        Wanda-TTPD & C4 + ETruthQA & 7.55 & 0.7582 & 0.9735 & 0.7332 & 0.8604 & 0.6868 & 0.8760$\pm$0.0039 \\
        OWL-TTPD & C4 & 7.69 & 0.541 & 0.9874 & 0.7337 & 0.8556 & 0.7104 & 0.8841$\pm$0.0045 \\
        OWL-TTPD & C4 + ETruthQA & 7.78 &  0.7572 & 0.9867 & 0.7543 & 0.8520 & 0.7158 & 0.8887$\pm$0.0025 \\
        SWL-TTPD & C4 & 7.81 & 0.6396 & 0.9542 & 0.7178 & 0.8561 &  0.6777 & 0.8579$\pm$0.0036 \\
        SWL-TTPD & C4 + ETruthQA & 7.62 & 0.7069 & 0.9595 & 0.7116 & 0.8606 & 0.6885 & 0.8692$\pm$0.0043 \\
        \textbf{TPLO-TTPD} & C4 & 7.65 & 0.7531 & 0.9901 & 0.7601 & \textbf{0.8774} & 0.7157 & 0.8716$\pm$0.0021 \\
        \textbf{TPLO-TTPD} & C4 + ETruthQA & 7.71 & \textbf{0.7691} & \textbf{0.9951} & \textbf{0.7797} & 0.8674 & \textbf{0.7204} & \textbf{0.8846$\pm$0.0037} \\
        \bottomrule
    \end{tabular}
    \caption{The experimental results on the True-False dataset using LLaMA2-13B-Chat. "Average" means average probing accuracies on 12 True-False datasets ("cities", "neg\_cities", "sp\_en\_trans", "neg\_sp\_en\_trans", "inventors", "neg\_inventors", "animal\_class", "neg\_animal\_class", "element\_symb", "neg\_element\_symb", "facts", "neg\_facts").}
    \label{LLaMA2-13B-Chat_results}
\end{table*}

\begin{table*}[h]
    \centering
    \small 
    \setlength{\tabcolsep}{3pt} 
    \renewcommand{\arraystretch}{0.8} 
    \begin{tabular}{l|c|c|cccccc}
        \toprule
        \textbf{Methods} &  \textbf{Calibration Data} & \textbf{Perplexity $\downarrow$} &
        \textbf{Cities} & \textbf{Neg\_Cities} & 
        \textbf{Invent.} & 
        \textbf{Facts} & \textbf{Neg\_Facts} & 
        \textbf{Average}\\
        \midrule
        Original-LR & N/A & 5.89 & 0.9911 & 0.9909 & 0.8826 & 0.8609 & 0.7977   & 0.9293$\pm$0.0107\\
        \midrule
        Wanda-LR  & C4 & 7.36 & 0.9251 & 0.8344 & 0.7710 & 0.7901 & 0.7423  & 0.8458$\pm$0.0154\\
        Wanda-LR & C4 + ETruthQA & 7.23 & 0.9335 & 0.8096 & 0.7645 & 0.7785 & 0.7304 & 0.8417$\pm$0.0190\\
        OWL-LR & C4 & 7.26 & 0.9476 & 0.8710 & 0.7779 & 0.7958 & 0.7475 & 0.8478$\pm$0.0124\\
        OWL-LR & C4 + ETruthQA & 7.35 & 0.9483 & 0.9022 & 0.7849 & 0.8028 & 0.7540 & 0.8551$\pm$0.0135\\
        SWL-LR & C4 & 7.41 & 0.9360 & 0.8602 & 0.7766 & 0.7844 & 0.7156 & 0.8198$\pm$0.0196 \\
        SWL-LR & C4 + ETruthQA & 7.45  & 0.9445 & 0.8990 & 0.7709 & 0.7918 & 0.7041 & 0.8254$\pm$0.0167 \\
        \textbf{TPLO-LR} & C4 & 7.32 & \textbf{0.9545} & 0.9059 & \textbf{0.7968}  & 0.8013 & \textbf{0.7653} & 0.8604$\pm$0.0031\\
        \textbf{TPLO-LR} & C4 + ETruthQA & 7.38 & 0.9541 & \textbf{0.9160} & 0.7901 & \textbf{0.8081} & 0.7612 & \textbf{0.8647$\pm$0.0161}\\
        \midrule
        Original-CCS & N/A & 5.89 & 0.7738 & 0.7806 & 0.6766 & 0.7707 & 0.7261  & 0.7973$\pm$0.0633\\
        \midrule
        Wanda-CCS & C4 & 7.36 & 0.6832 & 0.7033 & 0.5199 & 0.6636 & 0.6112 & 0.6151$\pm$0.0460\\
        Wanda-CCS & C4 + ETruthQA & 7.23 & 0.6553 & 0.6855 & 0.5656 & 0.5912 & 0.5391 & 0.6327$\pm$0.0645\\
        OWL-CCS  & C4 & 7.26 & 0.7807 & 0.7767 & 0.6080 & 0.6353 & 0.5786   & 0.6785$\pm$0.0669\\
        OWL-CCS  & C4 + ETruthQA & 7.35 & 0.7735 & 0.7641 & 0.6014 & 0.6721 & 0.6254 & 0.6875$\pm$0.0756 \\
        SWL-CCS & C4 & 7.41  & 0.6598 & 0.6555 & 0.5045 & 0.5963 & 0.5608 & 0.5824$\pm$0.0509\\
        SWL-CCS & C4 + ETruthQA & 7.45  & 0.6475 & 0.6257 & 0.5381 & 0.5806 & 0.5256 & 0.5736$\pm$0.0424\\
        \textbf{TPLO-CCS} & C4 & 7.32  & \textbf{0.8001} & 0.7901 & 0.6201 & 0.7313 & 0.6651 & 0.6901$\pm$0.0312\\
        \textbf{TPLO-CCS} & C4 + ETruthQA & 7.38 & 0.7909 & \textbf{0.7925} & \textbf{0.6256}  & \textbf{0.7423} & \textbf{0.6701} & \textbf{0.6995$\pm$0.0733}\\
        \midrule
        Original-MM & N/A & 5.89 & 0.9421 & 0.9957 & 0.7748 & 0.8612 &  0.7051 & 0.9051$\pm$0.0037 \\
        \midrule
        Wanda-MM & C4 & 7.36 & 0.8436 & 0.9554 & 0.6902 & 0.8315 & 0.6807 & 0.8316$\pm$0.0141\\
        Wanda-MM & C4 + ETruthQA & 7.23 & 0.8617 & 0.9555 & 0.7285 & 0.8377 & 0.6913 & 0.8421$\pm$0.0108\\
        OWL-MM & C4 & 7.26 & 0.8951 & 0.9624 & 0.7322 & 0.8445 & 0.6825 &0.8647$\pm$0.0093 \\
        OWL-MM & C4 + ETruthQA & 7.35 & 0.8918 & 0.9719 & 0.7011 & 0.8479 & 0.6780 & 0.8699$\pm$0.0102\\
        SWL-MM & C4 & 7.41  & 0.8164 & 0.8792 & 0.6733 & 0.8195 & 0.6611 & 0.8005$\pm$0.0142 \\
        SWL-MM & C4 + ETruthQA & 7.45 & 0.7963 & 0.8694 & 0.6878  & 0.8311 & 0.6742 & 0.8009$\pm$0.0129 \\
        \textbf{TPLO-MM} & C4 & 7.32  & 0.8998 & \textbf{0.9801} & 0.7401 & 0.8501 & 0.6976  & 0.8651$\pm$0.0013 \\
        \textbf{TPLO-MM} & C4 + ETruthQA & 7.38 & \textbf{0.9001} & 0.9787 & \textbf{0.7432} & \textbf{0.8559} & \textbf{0.7076} & \textbf{0.8702$\pm$0.0120}\\
        \midrule
        Original-TTPD & N/A & 5.89 & 0.9786 & 0.9875 & 0.8923 & 0.8774 & 0.7393  &0.9311$\pm$0.0029 \\
        \midrule
        Wanda-TTPD & C4 & 7.36 & 0.9230 & 0.9652 & 0.6860 & 0.8286 & 0.7127 & 0.8705$\pm$0.0032\\
        Wanda-TTPD & C4 + ETruthQA & 7.23 & 0.9271 & 0.9602 & 0.7195 & 0.8373 & 0.7227 & 0.8722$\pm$0.0038 \\
        OWL-TTPD & C4 & 7.26 & 0.9376 & 0.9767 & 0.7539 & 0.8552 & 0.7289 & 0.8825$\pm$0.0043 \\
        OWL-TTPD & C4 + ETruthQA & 7.35 & 0.9281 & 0.9739 & 0.7513 & 0.8529 & 0.7301 & 0.8831$\pm$0.0036 \\
        SWL-TTPD & C4 & 7.41 & 0.9070 & 0.9551 & 0.6672 & 0.8184 & 0.7090 & 0.8495$\pm$0.0035 \\
        SWL-TTPD & C4 + ETruthQA & 7.45  & 0.8908 & 0.9568 &  0.6806 & 0.8255 & 0.7028 & 0.8477$\pm$0.0040 \\
        \textbf{TPLO-TTPD} & C4 & 7.32 & 0.9354 & 0.9731 & \textbf{0.7801} & 0.8601 & 0.7321 & 0.8874$\pm$0.0012 \\
        \textbf{TPLO-TTPD} & C4 + ETruthQA & 7.38  & \textbf{0.9414} & \textbf{0.9790} & 0.7766 & \textbf{0.8676} & \textbf{0.7453} & \textbf{0.8915$\pm$0.0021} \\
        \bottomrule
    \end{tabular}
    \caption{The experimental results on the True-False dataset using Mistral-7B-Instruct-v0.3. "Average" means average probing accuracies on 12 True-False datasets ("cities", "neg\_cities", "sp\_en\_trans", "neg\_sp\_en\_trans", "inventors", "neg\_inventors", "animal\_class", "neg\_animal\_class", "element\_symb", "neg\_element\_symb", "facts", "neg\_facts").}
    \label{Mistral-7B-Instruct-v0.3_results}
\end{table*}

\begin{table*}[t]
\centering
\small
\setlength{\tabcolsep}{4pt}
\renewcommand{\arraystretch}{1.2}
\begin{tabular}{c|cc|ccc|ccccc}
\toprule
\textbf{Models} & \textbf{Methods} & \textbf{Calibration Data}  & \multicolumn{3}{c|}{\textbf{MC}}  & \multicolumn{3}{c}{\textbf{Open-Ended Generation}} \\
& & & \textbf{MC1$\uparrow$} & \textbf{MC2$\uparrow$} & \textbf{MC3$\uparrow$} & \textbf{\%Truth$\uparrow$} & \textbf{\%Info$\uparrow$} & \textbf{\%T*I$\uparrow$} \\
\midrule
LLaMA2-13B-Chat & Original & N/A  & 33.54 & 52.14 & 25.22  & 67.84 & 57.47 & 38.98\\
+ DoLa & Original & N/A &  35.19 & 64.37 & 32.05 & 68.25 & 58.62 & 40.01 \\
\midrule
LLaMA2-13B-Chat & Wanda & C4 & 24.13  & 42.27  & 17.11 & 60.13  & 50.25 & 30.21 \\
+ DoLa & Wanda &C4 & 26.76 & 44.51 & 24.31 & 61.58 & 51.63 & 31.79 \\
\midrule
LLaMA2-13B-Chat & Wanda & C4 + ETruthQA &  25.21 & 43.34 & 18.36 & 61.11  & 51.38 & 31.39  \\
+ DoLa & Wanda &C4 + ETruthQA & 27.45  & 45.71 & 25.89  & 62.25 & 52.65 & 32.77  \\
\midrule
LLaMA2-13B-Chat & OWL &C4 & 24.01  & 42.12 & 17.05 & 60.14  & 50.01 & 30.07 \\
+ DoLa  & OWL & C4 & 26.67 & 44.54 & 24.78  & 61.65 & 51.78 & 31.92 \\
\midrule
LLaMA2-13B-Chat & OWL & C4 + ETruthQA & 25.11  & 43.56 & 18.51 & 61.03  & 51.16 & 31.22 \\
+ DoLa  & OWL & C4 + ETruthQA & 27.65 & 45.78 & 25.85 & 62.56 & 52.78 & 33.02 \\
\midrule
LLaMA2-13B-Chat & TPLO (Ours) & C4  & 24.57  & 43.78 & 18.89 & 60.69 & 51.87 & 31.48 \\
+ DoLa  & TPLO (Ours)  & C4 & 26.91  & 46.01 & 27.32  & 61.88 & 52.91 & 32.74 \\
\midrule
LLaMA2-13B-Chat & TPLO (Ours) & C4 + ETruthQA & 25.78  & 44.81 & 19.91 & 61.52 & 52.35 & 32.21 \\
+ DoLa  & TPLO (Ours)  & C4 + ETruthQA & \textbf{27.81} & \textbf{46.93}  & \textbf{26.54}  & \textbf{62.91} & \textbf{53.65} & \textbf{33.75} \\
\bottomrule
\end{tabular}
\caption{Experimental results on TruthfulQA \cite{truthful_qa}: 1) multiple choice tasks (MC1, MC2, and MC3); and 2) open-ended generation tasks, where \%T*I stands for \%Truth*\%Info. We could see that pruning at 50\% sparsity  will degrade LLMs' performance on TruthfulQA and utilizing DoLa \cite{dola} can mitigate this degradation.}
\label{tab:truthqa_results_on_llama2_13b_chat}
\end{table*}

\begin{table*}[t]
\centering
\small
\setlength{\tabcolsep}{4pt}
\renewcommand{\arraystretch}{1.2}
\begin{tabular}{c|cc|ccc|ccccc}
\toprule
\textbf{Models} & \textbf{Methods} & \textbf{Calibration Data}  & \multicolumn{3}{c|}{\textbf{MC}}  & \multicolumn{3}{c}{\textbf{Open-Ended Generation}} \\
& & & \textbf{MC1$\uparrow$} & \textbf{MC2$\uparrow$} & \textbf{MC3$\uparrow$} & \textbf{\%Truth$\uparrow$} & \textbf{\%Info$\uparrow$} & \textbf{\%T*I$\uparrow$} \\
\midrule
LLaMA3.1-8B-Instruct & Original & N/A &  38.61 & 58.70 & 30.45  & 60.11 & 27.46 & 16.51 \\
+ DoLa & Original & N/A &  37.08 & 66.48 & 34.83 & 64.05 & 37.59 & 24.07 \\
\midrule
LLaMA3.1-8B-Instruct & Wanda & C4 & 29.18  & 48.81 & 22.77 & 55.13 & 23.65 & 13.03 \\
+ DoLa & Wanda & C4  & 30.13 & 56.24 & 30.15  & 59.21 & 33.67 & 19.93 \\
\midrule
LLaMA3.1-8B-Instruct & Wanda & C4 + ETruthQA & 30.22  & 50.01 & 23.95 & 56.41 & 24.21 & 13.66 \\
+ DoLa & Wanda & C4 + ETruthQA & 30.01 & 58.14 & 31.27  & 60.31 & 34.55 & 20.84 \\
\midrule
LLaMA3.1-8B-Instruct & OWL & C4 & 29.01 & 48.13 & 22.24 & 55.01 & 23.25 & 12.79  \\
+ DoLa  & OWL & C4 & 30.15  & 56.24 & 30.67 & 59.13 & 33.02 & 19.52 \\
\midrule
LLaMA3.1-8B-Instruct & OWL & C4 + ETruthQA & 30.11 & 50.01 & 24.01 & 56.12 & 24.09 & 13.52  \\
+ DoLa  & OWL & C4 + ETruthQA &  30.98 & 58.32 & 32.36 & 60.03 & 34.21 & 20.54 \\
\midrule
LLaMA3.1-8B-Instruct & TPLO & C4  & 29.49 & 49.55 & 23.63 & 57.01 & 24.89 & 14.19 \\
+ DoLa & TPLO & C4 & 29.01 & 57.31 & 30.89  & 60.21 & 34.41 & 21.06 \\
\midrule
LLaMA3.1-8B-Instruct & TPLO & C4 + ETruthQA & 30.81 & 50.74 & 24.95 & 57.89 & 25.42 & 14.72 \\
+ DoLa & TPLO & C4 + ETruthQA & \textbf{32.15} & \textbf{58.17} & \textbf{32.35} & \textbf{61.52} & \textbf{35.91} & \textbf{21.66} \\
\bottomrule
\end{tabular}
\caption{Experimental results on TruthfulQA \cite{truthful_qa}: 1) multiple choice tasks (MC1, MC2, and MC3); and 2) open-ended generation tasks, where \%T*I stands for \%Truth*\%Info. We could see that pruning at 50\% sparsity will degrade LLMs' performance on TruthfulQA and utilizing DoLa \cite{dola} can mitigate this degradation.}
\label{tab:truthqa_results_on_llama3.1_8b_instruct}
\end{table*}

\begin{table*}[t]
\centering
\small
\setlength{\tabcolsep}{4pt}
\renewcommand{\arraystretch}{1.2}
\begin{tabular}{c|c|cc}
\toprule
\textbf{Models} & \textbf{Perplexity} & \textbf{Probe on Cities} &  \textbf{Probe on Neg\_Cities}  \\
\midrule
Original & 6.10 & 0.9951 &  0.9986 \\
\midrule
Wanda of 0.4 sparsity & 7.01 & 0.8825 & 0.9701 \\
TPLO of 0.4 sparsity & 7.12 & 0.9213 & 0.9723 \\
\midrule
Wanda of 0.5 sparsity & 7.50 & 0.8439 & 0.9680 \\
TPLO of 0.5 sparsity & 7.65 & 0.8871 & 0.9765 \\
\midrule
Wanda of 0.6 sparsity & 8.21 & 0.8124 & 0.9011  \\
TPLO of 0.6 sparsity & 8.13 & 0.8321 & 0.9139 \\
\midrule
Wanda of 0.65 sparsity & 9.32 & 0.7631 & 0.8347   \\
TPLO of 0.65 sparsity  & 9.25 & 0.7912 & 0.8531  \\
\bottomrule
\end{tabular}
\caption{Experimental results on the relationship between lie detection accuracy (across city topics) and perplexity for LLaMA2-13B-Chat (LR, C4). As shown, TPLO achieves comparable perplexity to Wanda, while offering improved lie detection capabilities.}
\label{tab:tradeoff_on_llama2_13b_chat}
\end{table*}

\begin{table*}[t]
\centering
\small
\setlength{\tabcolsep}{4pt}
\renewcommand{\arraystretch}{1.2}
\begin{tabular}{c|c|cc}
\toprule
\textbf{Models} & \textbf{Perplexity} & \textbf{Probe on Cities} &  \textbf{Probe on Neg\_Cities}  \\
\midrule
Original & 8.28 & 0.9892 &  0.9942 \\
\midrule
Wanda of 0.4 sparsity & 10.14 & 0.9013 & 0.7325  \\
TPLO of 0.4 sparsity & 10.21 & 0.9431 & 0.7851 \\
\midrule
Wanda of 0.5 sparsity & 11.97 & 0.8968  & 0.7215 \\
TPLO of 0.5 sparsity & 11.91 & 0.9254 & 0.7661 \\
\midrule
Wanda of 0.6 sparsity & 12.56 & 0.8541 & 0.6881  \\
TPLO of 0.6 sparsity & 12.32 & 0.8951 & 0.7221 \\
\midrule
Wanda of 0.65 sparsity & 14.21 & 0.8154 & 0.6321   \\
TPLO of 0.65 sparsity  & 14.25 & 0.8431 & 0.6657  \\
\bottomrule
\end{tabular}
\caption{Experimental results on the relationship between lie detection accuracy (across city topics) and perplexity for LLaMA3.1-8B-Instruct (LR, C4). As shown, TPLO achieves comparable perplexity to Wanda, while offering improved lie detection capabilities.}
\label{tab:tradeoff_on_llama3.1_8b_instruct}
\end{table*}

\end{document}